\documentclass[letterpaper]{article} 
\usepackage{aaai24}  
\usepackage{times}  
\usepackage{helvet}  
\usepackage{courier}  
\usepackage[hyphens]{url}  
\usepackage{graphicx} 
\usepackage{natbib}  
\usepackage{caption} 
\usepackage{algorithm}
\usepackage{algorithmic}
\usepackage{subfig}
\usepackage{amsthm} 
\usepackage{enumitem}
\usepackage{amsmath,amssymb,amsfonts}
\usepackage{bm}
\usepackage{eqparbox}
\usepackage{graphicx}
\usepackage{xspace}
\usepackage{comment}
\usepackage{multirow}
\usepackage{arydshln}
\usepackage{tabularx}
\usepackage{tabulary}
\usepackage{booktabs}
\usepackage{makecell}

\usepackage{xspace}
\usepackage{circledsteps}

\usepackage{babel}
\usepackage[font=small,labelfont=bf]{caption}

\usepackage[hang,flushmargin]{footmisc}
\theoremstyle{definition}
\newtheorem{definition}{Definition}
\newtheorem{property}{Property}

\newcommand{\cmmnt}[1]{\ignorespaces} 

\newcommand{\argmax}{\operatornamewithlimits{argmax}}

\newcommand{\algname}{{DropTop}}

\makeatletter
\newcommand*\bigcdot{\mathpalette\bigcdot@{.5}}
\newcommand*\bigcdot@[2]{\mathbin{\vcenter{\hbox{\scalebox{#2}{$\m@th#1\bullet$}}}}}
\makeatother
\usepackage{newfloat}
\usepackage{listings}
\DeclareCaptionStyle{ruled}{labelfont=normalfont,labelsep=colon,strut=off} 
\floatstyle{ruled}
\newfloat{listing}{tb}{lst}{}
\floatname{listing}{Listing}
\title{Adaptive Shortcut Debiasing for Online Continual Learning}
\author{Doyoung Kim, Dongmin Park, Yooju Shin, Jihwan Bang, Hwanjun Song, Jae-Gil Lee\thanks{Jae-Gil Lee is the corresponding author.}}
\affiliations{
    KAIST, Daejeon, Republic of Korea\\ 
    {\{dodokim, dongminpark, yooju.shin, jihwan.bang, songhwanjun, jaegil\}}@kaist.ac.kr
}

\usepackage{bibentry}

\begin{document}

\maketitle


\begin{abstract}
   We propose a novel framework \algname{} that suppresses the shortcut bias in online continual learning\,(OCL) while being adaptive to the varying degree of the shortcut bias incurred by continuously changing environment. 
By the observed high-attention property of the shortcut bias, highly-activated features are considered candidates for debiasing.
More importantly, resolving the limitation of the online environment where prior knowledge and auxiliary data are not ready, two novel techniques---feature map fusion and adaptive intensity shifting---enable us to automatically determine the appropriate level and proportion of the candidate shortcut features to be dropped.
Extensive experiments on five benchmark datasets demonstrate that, when combined with various OCL algorithms, \algname{} increases the average accuracy by up to 10.4\% and decreases the forgetting by up to 63.2\%. 

\end{abstract}

\section{Introduction}
\label{sec:intro}

Deep neural networks\,(DNNs) often rely on a strong correlation between peripheral features, which are usually easy-to-learn, and target labels during the learning process\,\cite{park2021taufe,scimeca2021shortcut}. Such peripheral features and learning bias are called \emph{shortcut features} and \emph{shortcut bias}\,\cite{geirhos2018imagenet, hendrycks2021natural, scimeca2021shortcut, shah2020pitfalls}. For example, DNNs may extract only shortcut features such as color, texture, and local or background cues; when classifying dogs and birds, only the legs\,(i.e., local cue) and sky\,(i.e., background cue) could be extracted if they are the easiest to distinguish between the two classes. The shortcut bias is an important problem in most computer vision themes such as image classification\,\cite{geirhos2020shortcut}. 

\emph{Online continual learning\,(OCL)} for image classification\footnote{Unless otherwise specified, OCL is involved with image classification in this paper.} maintains a DNN to classify images from an online stream of images and tasks, where an upcoming task may include new classes\,\cite{de2021continual, rolnick2019experience, bang2021rainbow, buzzega2020dark}. Due to the characteristics of the online environment, there is not abundant training data\,(i.e., images), and the computational and memory budgets are typically tight\,\cite{de2021continual}. This limited opportunity for learning exacerbates the shortcut bias in OCL, because DNNs tend to learn easy-to-learn features early on\,\cite{du2021towards}. Therefore, we take a step forward to investigate its adverse effects in OCL.

\begin{figure}[t!]
\centering 
\includegraphics[width=0.99\columnwidth]{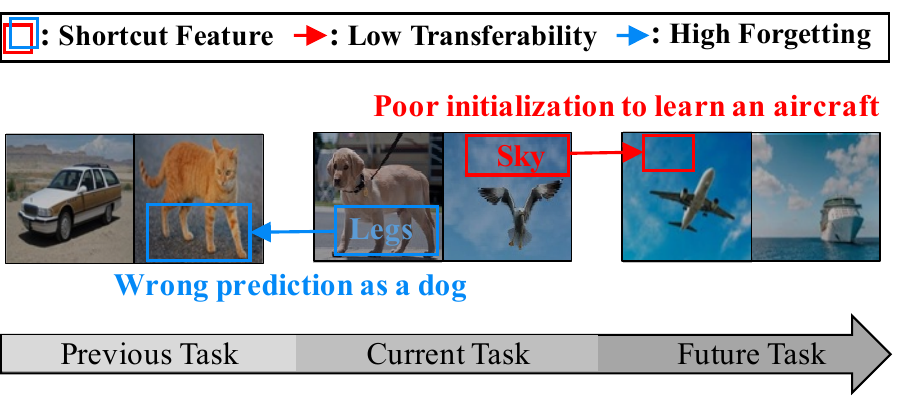}
\vspace*{-0.3cm}
\caption{\textbf{Negative effect of the shortcut bias in OCL}: shows the low transferability and high forgetability of shortcut features.}
\vspace*{-0.5cm}
\label{fig:neg_shortcutbias_ocl}
\end{figure}

The shortcut bias hinders the main goal of OCL that solves the plasticity and stability dilemma for \emph{high transferability} and \emph{low catastrophic forgetting}. That is, it incurs \emph{low} transferability and \emph{high} forgetting in OCL, because shortcut features do not generalize well to unseen new classes\,\cite{geirhos2020shortcut, park2021taufe} and are no longer discriminative for all classes. Figure \ref{fig:neg_shortcutbias_ocl} illustrates the negative effect of the shortcut bias in OCL. Regarding low transferability, if an OCL model learns to use the sky as a shortcut feature for the current task, it is faced with a bad initial point for the future task. Regarding high forgetting, if an OCL model learns to use the dog's legs for the current task, it is forced to forget the prior knowledge about the legs due to the misclassification of the animals other than the dog.

A significant amount of research has gone into eliminating an undesirable\,(e.g., shortcut) bias in \emph{offline} supervised learning. Representative debiasing methods require the prior knowledge of a target task to predefine the undesirable bias for unseen conditions\,\cite{bahng2020learning, geirhos2018imagenet, lee2019srm} or leverage auxiliary data such as out-of-distribution\,(OOD) data\,\cite{lee2021removing, park2021taufe}. However, \emph{neither} prior knowledge \emph{nor} auxiliary data is available in OCL, because future tasks are inherently unknown and access to auxiliary data is typically constrained due to a limited memory budget. Therefore, overcoming the lack of prior knowledge and auxiliary data should be the main challenge in debiasing for OCL. 

\begin{figure}[t!]
\centering
\includegraphics[width=0.99\columnwidth]{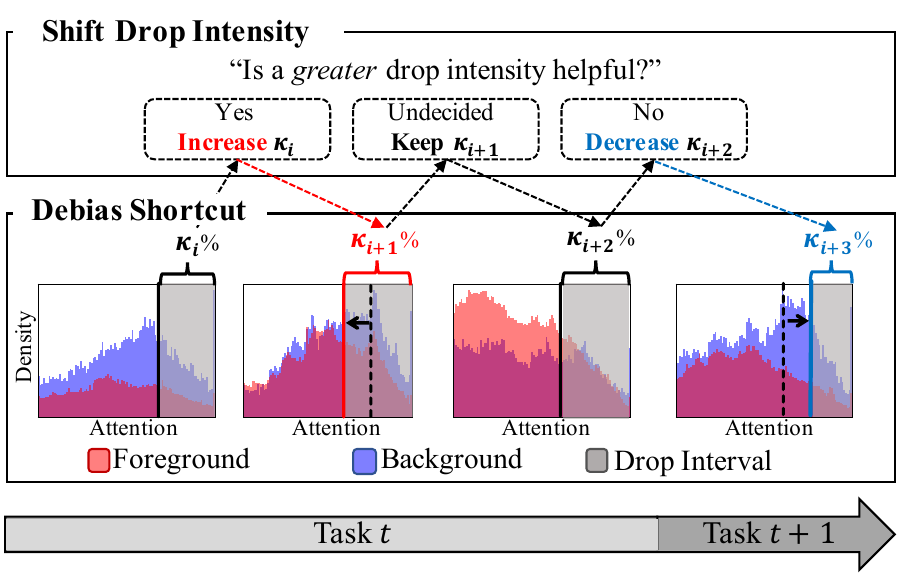}
\vspace*{-0.3cm}
\caption{\textbf{Adaptive intensity shifting}: shows how the portion of the high-attention features to be dropped can be adjusted in Split ImageNet, where the ground-truth\protect\footnotemark~background (shortcut) annotations are available for evaluation.} 
\label{fig:debiasing_shortcutbias_ocl}
\vspace*{-0.4cm}
\end{figure}

\footnotetext{The ground-truth information is used for generating the histograms for exposition purposes, but it is \emph{not} used in \algname{}.}

For shortcut debiasing, we suppress the learning of the shortcut features in a DNN. Accordingly, those shortcut features need to be identified accurately and efficiently without help from prior knowledge and auxiliary data. To this end, we pose \emph{two} research questions.

\textbf{RQ1} ``Which level of features is the most useful for identifying shortcut features?'': A DNN consists of many layers, and each layer produces its own feature map. Due to the aforementioned uncertainty, we take into account both low-level features from a low-level layer and high-level features from a high-level layer, refining semantic high-level features with structural low-level features via \emph{feature map fusion}. Our experience indicates that picking up the first and last layers is sufficient for the fusion. Meanwhile, it is known that DNNs prefer to learn shortcut features \emph{early} owing to their simplicity\,\cite{valle-perez2018deep, shah2020pitfalls, scimeca2021shortcut}, so they tend to have \emph{high attention scores}. Therefore, we drop the features with top-$\kappa\%$ attention scores on the \emph{fused} feature map.

\textbf{RQ2} ``How much portion of high-attention features are expected to be shortcut features?'': Answering this question is very challenging because the high-attention features include both shortcut and non-shortcut features, though shortcut features are expected to be dominating. Nevertheless, we endeavor to propose a novel, practical solution called \emph{adaptive intensity shifting}. In Figure \ref{fig:debiasing_shortcutbias_ocl}, if the features with top-$\kappa\%$ attention scores include a large number of shortcut\,(e.g., background) features, it is better to increase $\kappa$ to drop shortcut features aggressively (see the 1st histogram); for the opposite case possibly resulted from effective debiasing, it is better to decrease $\kappa$ not to drop useful features (see the 3rd histogram). 

Although the idea is very intuitive, there is no ground-truth for shortcut features in practice. Instead, we examine the \emph{loss reduction} followed by two shift directions---increment and decrement. A bigger loss reduction can be accomplished if more of shortcut features are dropped. 
The benefit of this loss-based approach is that estimating the dominance of shortcut features in high-attention features does not require additional overhead.

Overall, answering the two questions, we develop a novel framework, \textbf{\textit{\algname{}}}, for suppressing shortcut features in OCL. It is model-agnostic and thus can be incorporated into any replay-based OCL method; \emph{two} widely-used backbones, a convolutional neural network\,(CNN)\,\cite{he2016deep} and a Transformer\,\cite{dosovitskiy2021an}, are adopted for the evaluation. The main contributions of the paper are summarized as follows:
\begin{itemize}[leftmargin=10pt,nosep]
\smallskip
\item To the best of our knowledge, this is the \emph{first} work to address the shortcut bias in OCL. Moreover, we theoretically analyze its negative effect in OCL.
\item We present feature map fusion and adaptive intensity shifting to get around the uncertainty caused by the absence of prior knowledge and auxiliary data. 
\item We empirically show that \algname{} improves the average accuracy and forgetting of seven representative OCL methods by up to 10.4\% and 63.2\%, respectively, on conventional benchmarks as well as newly-introduced benchmarks, namely Split ImageNet-OnlyFG and Split ImageNet-Stylized.
\end{itemize}
\section{Related Work}
\label{sec:related_works}
 
\subsubsection{Online Continual Learning}
Recent studies have mainly considered an online environment for more practical applications\,\cite{rolnick2019experience, buzzega2020dark, prabhu2020gdumb, bang2021rainbow, koh2022online, chaudhry2018riemannian, kirkpatrick2017overcoming}. Access to the data from the current task is permitted until model convergence in an offline environment, but only once in an online environment. Thus, online continual learning\,(OCL) is much more challenging than offline continual learning.
Representative OCL methods are mainly categorized into two groups: \emph{replay-based} strategies exploiting small episodic memory to consolidate the knowledge of old tasks\,\cite{rolnick2019experience, buzzega2020dark, prabhu2020gdumb, koh2022online} and \emph{regularization-based} strategies penalizing a model for rapid parameter updates to avoid the forgetting of old tasks\,\cite{chaudhry2018riemannian, kirkpatrick2017overcoming}. In general, replay-based approaches outperform regularization-based ones in terms of accuracy and computational efficiency. Thus, we focus on the replay-based approaches in this paper.

Replay-based approaches suggest the policies for keeping more useful instances. ER\,\cite{rolnick2019experience} and DER++\,\cite{buzzega2020dark} propose random sampling and reservoir sampling, respectively. DER++\,\cite{buzzega2020dark} additionally utilizes knowledge distillation\,\cite{hinton2015distilling} to better retain previous knowledge. MIR\,\cite{mir} selects the samples whose loss increases the most by a preliminary model update. GSS\,\cite{aljundi2019gradient} chooses the samples diversifying their gradients. Last, ASER\,\cite{shim2021online} retrieves the samples that better preserve latent decision boundaries for known classes while learning new classes.

\subsubsection{Debiasing Shorcut Features}
The negative impact of shortcuts bias has gained a great attention in the deep learning community\,\cite{ilyas2019adversarial, wang2020high}.
To prevent the overfitting to the shortcuts, several methods pre-define the type of the target shortcut bias. ReBias\,\cite{bahng2020learning} removes pixel-level local shortcuts using a set of biased predictions from a bias-characterizing model. Stylized-ImageNet\,\cite{geirhos2018imagenet} generates texture-debiased examples. SRM\,\cite{lee2019srm} synthesizes debiased examples via generative modeling. 
Another direction is to leverage auxiliary OOD data\,\cite{lee2021removing, park2021taufe}, which relies on undesirable features found in the OOD data.
However, these existing methods are not directly applicable to our environment, because enforcing prior knowledge or auxiliary data violates the philosophy of OCL.

Moreover, the adverse impact of shortcuts is amplified when training data is insufficient\,\cite{lee2021removing}---as in OCL involved with an online stream of small tasks. Overall, lack of the prerequisite and rich training data calls for a new debiasing method dedicated for OCL.
\section{Preliminary}
\label{sec:preliminary}

\begin{figure*}[!t]
    \centering
    \includegraphics[width=0.93\textwidth]{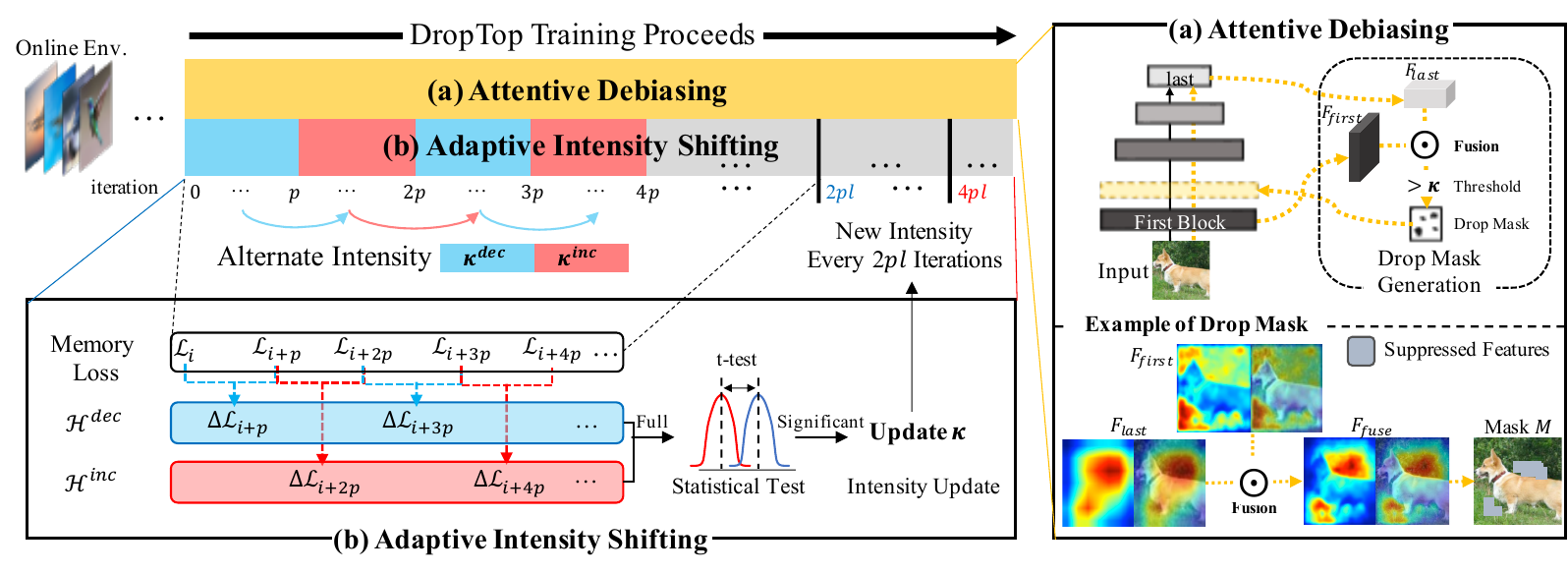}
    \vspace*{-0.4cm}
    \caption{\textbf{Detailed view of the main components of \algname{}}: (a) drops the highly-activated features on the fused feature map according to the intensity $\kappa$; as in the example of the drop mask, the fusion facilitates identifying the background shortcuts from the signal of the high-level features on important parts such as the body of the dog. (b) adjusts the drop intensity $\kappa$ for the use in attentive debiasing. 
    }
    \label{fig:model_arch}
    \vspace*{-0.6cm}
\end{figure*}

\subsubsection{Online Continual Learning}
We consider the online continual learning setting, where a sequence of tasks continually emerges with a set of new classes. Each data instance can be accessed only once unless it is stored in episodic memory. For the $t$-th task, let $D_t=\{(x^i_t,y^i_t)\}_{i=1}^{m_t}$ be a data stream obtained from a joint data distribution over $\mathcal{X}_t \times \mathcal{Y}_t$, where $\mathcal{X}_t$ is the input space, $\mathcal{Y}_t$ is the label space in a one-hot fashion, and $m_t$ is the number of instances of the $t$-th task. The goal of OCL is to train a classifier, such that it maximizes the test accuracy of all seen tasks $T=\{1,2, \dots, t\}$ at time $t$, with limited or no access to the data stream $D_{t^{\prime}}$ of previous tasks $t^{\prime}<t$.

\subsubsection{Shortcut and Non-shortcut Features}
Let a DNN model trained on $(x,y) \sim \mathcal{X}\times\mathcal{Y}$ comprises a general feature extractor $f^{\theta}:\mathcal{X}\rightarrow\mathcal{Z}\in\mathbb{R}^d$ and a final classifier layer $g^{\text{w}}:\mathcal{Z}\rightarrow \mathbb{R}^{k}$, which is the multiplication of a linear weight matrix $\text{w}\in\mathbb{R}^{k\times d}$ and a feature $z\in \mathcal{Z}$ such that $g^{\text{w}}(z)=w\cdot z$. Here, $k$ is the number of classes, and $d$ is the dimensionality of a feature $z$. Let a \emph{feature} be the function mapping from the input space $\mathcal{X}$ to a real number; formally speaking, for the $i$-th sub-feature extractor of $f^{\theta}$ where $i\!\in\!\{1,2,\!\ldots\!,d\}$, $f^{\theta}_i:\mathcal{X}\rightarrow\mathbb{R}$. DNNs are known to have simplicity bias toward shortcut features which can easily distinguish between given classes\,\cite{geirhos2018imagenet, shah2020pitfalls, geirhos2020shortcut, pezeshki2021gradient}. Then, the \emph{shortcut} and \emph{non-shortcut} features are defined by Definitions \ref{def:shortcut_feature} and \ref{def:non_shortcut_feature}, respectively.

\begin{definition}{\sc (Shortcut Feature)}.
Let $x$ and $\tilde{x}$ be the instances from seen and unseen data distributions $\mathcal{X}$ and $\mathcal{\tilde{X}}$, respectively, where $\mathcal{X}\cap\mathcal{\tilde{X}}=\emptyset$. We call a feature a \emph{shortcut} if it is not only highly activated ($\geq \rho$) for the seen data instance but also undesirably activated ($\geq \epsilon$) for the unseen data instance in expectation. That is,
\vspace{-0.1cm}
\begin{equation}
f^{\theta}_{s}(\rho, \epsilon) : \mathbb{E}_{x \sim \mathcal{X}} \big[|f^{\theta}_s(x)| \big] \geq \rho ~\land ~ \mathbb{E}_{\tilde{x} \sim \mathcal{\tilde{X}}} \big[|f^{\theta}_s(\tilde{x})| \big] \geq \epsilon,
\label{eq:shortcut_feature}
\vspace{-0.1cm}
\end{equation}
where $s$ is the index of the shortcut feature. $\rho$ and $\epsilon$ are the thresholds indicating high activation for the seen and unseen data distributions. 
\label{def:shortcut_feature}
\vspace{-0.1cm}
\end{definition}

Note that a shortcut feature, such as an animal's legs, can be observed from the unseen data instances because it is not an intrinsic feature.

\begin{definition}{\sc (Non-shortcut Feature)}.
Let $x$ and $\tilde{x}$ be the instances from seen and unseen data distributions $\mathcal{X}$ and $\mathcal{\tilde{X}}$, respectively, where $\mathcal{X}\cap\mathcal{\tilde{X}}=\emptyset$. We call a feature a \emph{non-shortcut} if it is highly activated only for the seen instance in expectation. That is,
\vspace{-0.1cm}
\begin{equation}
f^{\theta}_{n}(\rho, \epsilon) : \mathbb{E}_{x \sim \mathcal{X}} \big[|f^{\theta}_n(x)| \big] \geq \rho ~\land ~ \mathbb{E}_{\tilde{x} \sim \mathcal{\tilde{X}}} \big[|f^{\theta}_n(\tilde{x})| \big] < \epsilon. 
\label{eq:non_shortcut_feature}
\vspace{-0.1cm}
\end{equation}
\label{def:non_shortcut_feature}
\vspace{-0.5cm}
\end{definition}

\subsubsection{Property of Shortcut Feature} Due to a DNN's simplicity bias, the shortcut features tend to have higher activation values than the non-shortcut features\,\cite{shah2020pitfalls, pezeshki2021gradient}, such that
\vspace{-0.1cm}
\begin{equation}
\mathbb{E}_{s\in \mathcal{S}} \big[|f^{\theta}_s(x)| \big]> \mathbb{E}_{n\in \mathcal{N}} \big[|f^{\theta}_n(x)| \big],
\label{eq:shortcut_feature_property}
\vspace{-0.1cm}
\end{equation}
where $\mathcal{S}$ and $\mathcal{N}$ are the set of the indices of shortcut and non-shortcut features, respectively.


\section{Proposed Method: \algname{}}
\label{sec:methodology}

\algname{} realizes adaptive feature suppression thorough (a) \emph{attentive debiasing} with \emph{feature map fusion} and (b) \emph{adaptive intensity shifting}. The former drops high-attention features which are expected to be shortcuts, following the guidance from the latter in the form an adjusted drop intensity. Figure \ref{fig:model_arch} illustrates the detailed procedure of \algname{}. While (a) attentive debiasing is executed during the learning process, the drop intensity $\kappa$ is periodically adjusted by (b) adaptive intensity shifting. \algname{} can be implemented on top of any replay-based OCL methods, and see the implementation detail in Appendix A.

\subsection{Attentive Debiasing}
\label{subsec: Attentive Debiasing Layer}

The shortcut features generally exhibit higher attention scores than the non-shortcut feature by Eq.\ \eqref{eq:shortcut_feature_property}. This shortcut bias usually appears as local or background cues\,\cite{scimeca2021shortcut}. Thus, after fusing the feature maps of different levels, we generate a \textrm{drop mask} to debias the effect of local or background cues on the feature activation.

\subsubsection{Feature Map Fusion}
The high-level features are essential to obtain discriminative signals, but they lack fine-grained details\,\cite{zhong2019shallow,wei2021shallow}, e.g., boundaries of objects. The low-level features embrace structural information with more details, which is helpful for accurately finding shortcut regions as empirically shown in Section Ablation Study. Thus, feature map fusion facilitates the reduction of the ambiguity in identifying the shortcut regions. Specifically, as shown in Figure \ref{fig:model_arch}(a), given a small memory footprint, we fuse the feature maps ${F}_{first}\in\mathbb{R}^{h\times w\times c}$ and ${F}_{last}$ $\in\mathbb{R}^{h^{\prime}\times w^{\prime}\times c^{\prime}}$ respectively from the first and last layers. To cope with different resolutions, i.e., $\langle {h, w, c}\rangle \neq \langle {h^{\prime}, w^{\prime}, c^{\prime} }\rangle$, the last layer's feature map $F_{last}$ is up-sampled to have the same resolution as the first layer's feature map $F_{first}$ such that $h^{\prime}=h$ and $w^{\prime}=w$. Then, we conduct average pooling along the channel dimension, compressing the outputs of all $c$ channels to produce the attention map ${A}_{fuse}$, which is derived by
\vspace*{-0.1cm}
\begin{equation}
\begin{gathered}
{F}_{fuse} = {F}_{first}\odot {\rm Upsample}({F}_{last})\in\mathbb{R}^{h\times w \times c}~~\text{and}\\
{A}_{fuse} = {\rm ChannelPool}(F_{fuse})\in\mathbb{R}^{h\times w},
\end{gathered}
\label{eq:feature_fusion}
\vspace*{-0.1cm}
\end{equation}
where {\rm ChannelPool($\cdot$)} is the pooling along the channel, and $\odot$ is the element-wise tensor product.

\subsubsection{Drop Mask Generation}
Given a drop intensity $\kappa\%$ for each class\footnote{The class is used only during the training process for debiasing. We do not perform such feature suppression during testing.}, we generate a \textit{drop mask} $M \in \mathbb{R}^{h \times w}$ based on the attention map ${A}_{fuse}$ in Eq.\ \eqref{eq:feature_fusion}. The mask at $(i,j)$ in $M$ is determined by
\vspace*{-0.15cm}
\begin{equation}
M_{i,j} =
\begin{cases}
0 & \text{if}\!\!\quad (i,j) \in {\rm top\text{-}}\kappa(A_{fuse}) \\ 
1 & \text{otherwise},\\ 
\end{cases}
\label{eq:drop_mask}
\vspace*{-0.15cm}
\end{equation}
where ${\rm top\text{-}}\kappa(A_{fuse})$ returns the set of the top-$\kappa\%$ elements of $A_{fuse}$. While this \emph{hard} drop mask is simple and effective, we also test a \emph{soft} drop mask with continuous values in Appendix E. Last, we apply the drop mask to the first feature map ${F}_{first}$ after the stem layer of a backbone network, e.g., a ResNet or a Vision Transformer\,(ViT),
\vspace*{-0.1cm}
\begin{equation}
\Tilde{F}_{first} = M \odot {F}_{first}.
\label{eq:drop_ops}
\end{equation}
This masking process affects all succeeding layers along the forward process. The masked feature map $\Tilde{F}_{first}$ accomplishes debiasing by using only the attention scores which can be computed on the fly.

\subsection{Adaptive Intensity Shifting}
\label{subsec:AdaptiveIntensityShifting}

The extent of the shortcut bias naturally varies depending on the incoming tasks and the learning progress of a DNN model. Thus, \textit{adaptive intensity shifting} aims to guide {attentive debiasing} by adaptively adjusting the drop intensity $\kappa$ in a timely manner. The drop intensity is maintained separately for each class to capture diverse sensitivity to shortcut bias. (See Appendix E for an additional test of sharing the drop intensity across classes.) It is \emph{decreased} if the removal of features is expected to lose important non-shortcut features which have high predictive power, \emph{increased} if the removal of features is expected to help emphasize important non-shortcut features, and \emph{unchanged} in neither of the cases,
\vspace*{-0.15cm}
\begin{equation}
\kappa \leftarrow \begin{cases}
\kappa^\prime * \alpha & \text{if decrement} \\
\kappa^\prime * (1/\alpha) & \text{if increment} \\
\kappa^\prime & \text{otherwise},
\end{cases}
\label{eq:shifting}
\vspace*{-0.15cm}
\end{equation}
where $\kappa^\prime$ is the preceding value, and $\alpha$\,$(<1.0)$ is a hyperparameter that indicates the shift step.

\subsubsection{Loss Collection} 
To classify each case, we focus on the contribution to the reduction of the training loss because a better choice of the drop intensity leads to a better training performance\,\cite{scimeca2021shortcut, geirhos2020shortcut}. To determine whether a decrement or an increment is preferable, two potential values $\kappa=\kappa^\prime * \alpha$ and $\kappa=\kappa^\prime * (1/\alpha)$ alternate every $p$ iterations. $p$ is set to be long enough to observe \emph{stable} behavior with respect to each $\kappa$ option, as shown in Appendix D. The loss reductions are computed at the end of every $p$ iterations and maintained in the set $\mathcal{H}^{dec}$ when $\kappa^\prime * \alpha$ is used and in the set $\mathcal{H}^{inc}$ when $\kappa^\prime * (1/\alpha)$ is used, as shown in Figure \ref{fig:model_arch}(b),
\vspace*{-0.1cm}
\begin{equation}
\begin{gathered}
\tiny
\,\,\,{\mathcal{H}}^{dec} = \{\Delta\mathcal{L}_0,\Delta\mathcal{L}_{2p},\dots,\Delta\mathcal{L}_{2p(l-1)}\} \text{\small and} \\
{\mathcal{H}}^{inc} = \{\underbrace{\Delta\mathcal{L}_{p},\Delta\mathcal{L}_{3p},\dots,\Delta\mathcal{L}_{2pl}}_{\text{Collect  } \Delta\mathcal{L} \text{  every  } 2p \text{  iterations}}\},
\end{gathered}
\vspace*{-0.1cm}
\end{equation}
where $\Delta\mathcal{L}_{(q+1)\cdot p} = \mathcal{L}_{q\cdot p}-\mathcal{L}_{(q+1)\cdot p}$, and $\mathcal{L}_{(q+1) \cdot p}$ is the expected cross-entropy loss on the samples of a specific class in the memory buffer at the iteration $(q+1)\cdot p$. Note that we compute the loss using the samples in the memory buffer rather than in a batch to obtain a more generalizable training loss. Those loss reductions are recorded until ${\mathcal{H}}^{dec}$ and ${\mathcal{H}}^{inc}$ have $l$ elements. Only a single model is needed to compare two shift directions, preserving the memory constraint of OCL; the validity of using a single model is empirically confirmed in Section Evaluation.

\subsubsection{Statistical Test and Update}
When ${\mathcal{H}}^{dec}$ and ${\mathcal{H}}^{inc}$ become full every $2\cdot p \cdot l$ iterations, we conduct $t$-test to evaluate if the difference between the two directions in the loss reduction is statistically significant, i.e., either $p$-value $\leq 0.05$ or $p$-value $\geq 0.95$. If a better direction exists, the preceding value is updated to the better direction; otherwise, it remains the same. Appendix A describes the pseudocode of adaptive intensity shifting, which is self-explanatory.

\subsubsection{Training Data Stabilization}
Furthermore, we consider the side effect of shifting the drop intensity during training. If it changes over time, the distribution of input features could be swayed inconsistently. As widely claimed in \citeauthor{ioffe2015batch, zhou2022online}, this problem often causes training instability and eventually deteriorates the overall performance. In this sense, we decide to drop $\gamma\%$ of the features \emph{constantly}. Thus, $\kappa$ cannot grow beyond $\gamma$; if $\kappa < \gamma$, $(\gamma-\kappa)$\% of the features are additionally chosen uniformly at random among those whose drop mask is not $1$.

\subsection{Understanding of Shortcut Bias in OCL}
\label{sec:theoretical_analysis}

From a theoretical standpoint, we clarify two detrimental effects of learning shortcut features in OCL. See Appendix B for relevant empirical evidences. Let $g^{\text{w}}_i:\mathbb{R}\rightarrow\mathbb{R}^{k}$ be the multiplication of an $i$-th column vector of the weight matrix $\text{w}$ and an $i$-th feature $f_i^\theta$. Then, $g^{\text{w}}_i(f^{\theta}_i(x))$ denotes the contribution of the $i$-th feature to a prediction, i.e., how much the $i$-th feature $f^{\theta}_i(x)$ contributes to the prediction $g^{\text{w}}(f^{\theta}(x))$, since $g^{\text{w}}(f^{\theta}(x))= \sum_{i=1}^d{g^{\text{w}}_i(f^{\theta}_i(x))}$. 

\begin{property}[{\sc Low Transferability}]
A shortcut feature of the current task hinders the model from learning the knowledge of the next task.
\end{property}
\begin{proof}
\vspace*{-0.2cm}
Let's consider training a classifier for the ($t$+1)-th task, given a shortcut biased model $\theta_t$ as the initial model. By Eq.~\eqref{eq:shortcut_feature}, $\mathbb{E}_{x_{t+1} \sim \mathcal{X}_{t+1}}\big[|f^{\theta_t}_s(x_{t+1})| \big] \geq \epsilon$ if $x_{t+1}$ shares the same shortcut features with $x_t$. Then, when training a new model $\theta_{t+1}$ from the model $\theta_{t}$, the shortcut feature $f^{\theta_t}_s$ is no longer discriminative since it can appear in both $\mathcal{X}_{t}$ and $\mathcal{X}_{t+1}$. Thus, the model $\theta_{t+1}$ should learn a totally new discriminative feature, starting from a poor initialization. As a result, the training convergence becomes slower compared with a non-biased model $\theta_t^*$.
\end{proof}

\begin{property}[{\sc High Forgetability}]
A shortcut feature of the current task makes the model forget the knowledge of the previous tasks.
\end{property}
\begin{proof}
\vspace*{-0.2cm}
Given the previous tasks at $t^{\prime}\in\{1,\ldots,t-1\}$ and the current task at $t$, let's assume that we trained a shortcut biased model $\theta_t$.
Then, the shortcut feature $s$ of $f^{\theta_t}$ contributes to predict the class $y_t\in\mathcal{Y}_t$ such that $\argmax g^{\text{w}_t}_s(f^{\theta_t}_s(x_t))=y_t\in\mathcal{Y}_t$. However, by Eq.~\eqref{eq:shortcut_feature}, $\mathbb{E}_{x_{t^{\prime}} \sim \mathcal{X}_{t^{\prime}}}\big[|f^{\theta_t}_s(x_{t^{\prime}})| \big] \geq \epsilon$ if $x_{t^{\prime}}$ shares the same shortcut features with $x_{t}$.
Then, even for the previous instance $x_{t^{\prime}}$, the shortcut feature $s$ contributes to predict the class $y_{t}\in\mathcal{Y}_{t}$ such that $\argmax g^{\text{w}_t}_s(f^{\theta_t}_s(x_{t^{\prime}}))=y_t\notin\mathcal{Y}_{t^{\prime}}$, which is a wrong prediction.
Due to the high activation of shortcut features in Eq.~\eqref{eq:shortcut_feature_property}, this wrong feature contribution gives a significant impact to the final prediction $g^{\text{w}}(f^{\theta}(x_{t^{\prime}}))= \sum_{i=1}^d{g^{\text{w}}_i(f^{\theta}_i(x_{t^{\prime}}))}$, so that the model highly forgets the knowledge of the previous tasks. 
\end{proof}

\section{Evaluation}
\label{sec:evaluation}

\subsection{Experiment Setting}

\subsubsection{Dataset Preparation}
We validate the debiasing efficacy of \algname{} on \textit{biased} and \textit{unbiased} setups. A biased setup measures the OCL performance on biased test datasets, where the test dataset shares the same distribution of shortcut features with the training dataset, e.g., the sky background in the bird class. On the other hand, an unbiased setup\,\cite{bahng2020learning} does not contain the shortcut features in the test dataset, so that we can more clearly measure the debiasing efficacy.

\noindent{\underline{{Biased Setup:}}} We use the \emph{Split CIFAR-10}\,\cite{krizhevsky2009learning}, \emph{Split CIFAR-100}\,\cite{krizhevsky2009learning}, and \emph{Split ImageNet-9}\,\cite{xiao2020noise} for the biased setup. Split CIFAR-10 and Split CIFAR-100 consist of five different tasks with non-overlapping classes, and thus each task contains two and 20 classes, respectively. The details of Split ImageNet-9 are provided in Appendix C. 

\noindent{\underline{{Unbiased Setup:}}} The evaluation of unbiased setup is conducted after being trained on \textit{Split ImageNet-9}. We introduce two \textit{Split ImageNet-9} variants: \emph{Split ImageNet-OnlyFG} and \emph{Split ImageNet-Stylized}, which are generated by debiasing two realistic shortcuts, the background and the local texture cue, respectively. In ImageNet-OnlyFG\,\cite{xiao2020noise}, the background is removed to evaluate the dependecy on the background in image recognition; in ImageNet-Stylized\,\cite{geirhos2018imagenet}, the local texture is shifted by style-transfer, and the reliance of a model on the local texture cue is removed. Performance improvements on these two datasets indicate that \algname{} helps untie undesirable dependency on the realistic shortcuts. 

\subsubsection{Algorithms and Evaluation Metrics}
We apply \algname{} to seven popular algorithms including {ER}\,\cite{rolnick2019experience} and {DER++}\,\cite{buzzega2020dark}, {MIR}\,\cite{mir}, {GSS}\,\cite{aljundi2019gradient}, {ASER}\,\cite{shim2021online}, L2P\,\cite{wang2022learning}, and DualPrompt\,\cite{wang2022dualprompt}. 
They are all actively being cited. The experimental details are presented in Appendix E. 
We adopt widely-used performance measures\,\cite{aljundi2019gradient, shim2021online, koh2022online}: (1) \emph{average accuracy} $A_{avg} = \frac{1}{T}\Sigma_{i=1}^{T}A_i$, where $A_i$ is the accuracy at the end of the $i$-th task, and (2) \emph{forgetting} $F_{last}= \frac{1}{T-1}\Sigma_{j=1}^{T-1}f_{T,j}$, where $f_{i,j}$ indicates how much the model forgets about the $j$-th task after learning the $i$-th task ($j<i$). For reliability, we repeat every experiment \emph{five} times with different random seeds and report the average value with the standard error. We report the relative improvements (denoted as ``Rel.~Improv.'') by \algname{} for each metric. All algorithms are implemented using PyTorch 1.12.1 and tested on a single NVIDIA RTX 2080Ti GPU, and the source code is available at \url{https://github.com/kaist-dmlab/DropTop}.

\newcolumntype{C}[1]{>{\centering\arraybackslash}p{#1}}
\def\arraystretch{0.8}
\begin{table*}[t]
\small
\centering
\vspace*{-0.2cm}
\resizebox{1.0\linewidth}{!}{%
\begin{tabular}[c]
{@{}l||cccccc|ccccc@{}}
\toprule
\multicolumn{1}{c||}{} & \multicolumn{6}{c|}{\textbf{\underline{    Biased Setup    }}} 
& \multicolumn{4}{c}{\textbf{\underline{    Unbiased Setup   }}} \\ \addlinespace[0.6ex]
\multicolumn{1}{c||}{\textbf{}}& \multicolumn{2}{c}{\textbf{Split CIFAR-100}} & \multicolumn{2}{c}{\textbf{Split CIFAR-10}} & \multicolumn{2}{c|}{\textbf{Split ImageNet-9}}
& \multicolumn{2}{c}{\textbf{Split ImageNet-OnlyFG}} & \multicolumn{2}{c}{\textbf{Split ImageNet-Stylized}} \\
\multicolumn{1}{c||}{\textbf{Method}}
& \multicolumn{1}{c}{\bm{$A_{avg}$}\,($\uparrow$)}
& \multicolumn{1}{c|}{\bm{$F_{last}$}\,($\downarrow$)}
& \multicolumn{1}{c}{\bm{$A_{avg}$}\,($\uparrow$)}
& \multicolumn{1}{c|}{\bm{$F_{last}$}\,($\downarrow$)}
& \multicolumn{1}{c}{\bm{$A_{avg}$}\,($\uparrow$)}
& \multicolumn{1}{c|}{\bm{$F_{last}$}\,($\downarrow$)}
& \multicolumn{1}{c}{\bm{$A_{avg}$}\,($\uparrow$)}
& \multicolumn{1}{c|}{\bm{$F_{last}$}\,($\downarrow$)}
& \multicolumn{1}{c}{\bm{$A_{avg}$}\,($\uparrow$)}
& \multicolumn{1}{c}{\bm{$F_{last}$}\,($\downarrow$)}
\\ \toprule
\multirow{1}{*}[0.2em]{ER}                                
& \begin{tabular}[c]{@{}c@{}}{ {25.3}\,($\pm$0.2) }\end{tabular}
& \begin{tabular}[c]{@{}c@{}}{ {19.2}\,($\pm$0.3) }\end{tabular}
& \begin{tabular}[c]{@{}c@{}}{ {55.4}\,($\pm$0.3) }\end{tabular}
& \begin{tabular}[c]{@{}c@{}}{ {39.9}\,($\pm$2.4) }\end{tabular}
& \begin{tabular}[c]{@{}c@{}}{ {46.1}\,($\pm$0.2) }\end{tabular}
& \begin{tabular}[c]{@{}c@{}}{ {55.8}\,($\pm$1.8) }\end{tabular}
& \begin{tabular}[c]{@{}c@{}}{ {32.8}\,($\pm$0.4) }\end{tabular}
& \begin{tabular}[c]{@{}c@{}}{ \textbf{45.0}\,($\pm$1.5) }\end{tabular}
& \begin{tabular}[c]{@{}c@{}}{ {39.3}\,($\pm$0.3) }\end{tabular}
& \begin{tabular}[c]{@{}c@{}}{ {49.7}\,($\pm$0.9) }\end{tabular}

\\
\multirow{1}{*}[0.2em]{\textbf{~~$+$\algname{}}}  
& \begin{tabular}[c]{@{}c@{}}{ \textbf{26.5}\,($\pm$0.2) }\end{tabular}
& \begin{tabular}[c]{@{}c@{}}{ \textbf{18.3}\,($\pm$1.3) }\end{tabular}
& \begin{tabular}[c]{@{}c@{}}{ \textbf{61.0}\,($\pm$0.6) }\end{tabular}
& \begin{tabular}[c]{@{}c@{}}{ \textbf{37.6}\,($\pm$1.9) }\end{tabular}
& \begin{tabular}[c]{@{}c@{}}{ \textbf{48.9}\,($\pm$0.4) }\end{tabular}
& \begin{tabular}[c]{@{}c@{}}{ \textbf{44.0}\,($\pm$3.8) }\end{tabular}
& \begin{tabular}[c]{@{}c@{}}{ \textbf{35.9}\,($\pm$0.9) }\end{tabular}
& \begin{tabular}[c]{@{}c@{}}{ {47.8}\,($\pm$3.7) }\end{tabular}
& \begin{tabular}[c]{@{}c@{}}{ \textbf{41.3}\,($\pm$0.1) }\end{tabular}
& \begin{tabular}[c]{@{}c@{}}{ \textbf{38.9}\,($\pm$2.6) }\end{tabular}
\\ \addlinespace[0.3ex]\cdashline{1-11}\addlinespace[0.5ex]
\multirow{1}{*}[0.2em]{~~Rel. Improv.}  
& \begin{tabular}[c]{@{}c@{}}{ 4.4\% }\end{tabular}
& \begin{tabular}[c]{@{}c@{}}{ 4.9\% }\end{tabular}
& \begin{tabular}[c]{@{}c@{}}{ 10.2\% }\end{tabular}
& \begin{tabular}[c]{@{}c@{}}{ 5.7\% }\end{tabular}
& \begin{tabular}[c]{@{}c@{}}{ 6.1\% }\end{tabular}
& \begin{tabular}[c]{@{}c@{}}{ 21.1\% }\end{tabular}
& \begin{tabular}[c]{@{}c@{}}{ 9.4\% }\end{tabular}
& \begin{tabular}[c]{@{}c@{}}{ -6.2\% }\end{tabular}
& \begin{tabular}[c]{@{}c@{}}{ 5.1\% }\end{tabular}
& \begin{tabular}[c]{@{}c@{}}{ 21.8\% }\end{tabular}
\\
\midrule

\multirow{1}{*}[0.2em]{DER++}                                                         
& \begin{tabular}[c]{@{}c@{}}{ {23.6}\,($\pm$0.3) }\end{tabular}
& \begin{tabular}[c]{@{}c@{}}{ {33.4}\,($\pm$0.7) }\end{tabular}
& \begin{tabular}[c]{@{}c@{}}{ {59.8}\,($\pm$0.8) }\end{tabular}
& \begin{tabular}[c]{@{}c@{}}{ {28.2}\,($\pm$1.5) }\end{tabular}
& \begin{tabular}[c]{@{}c@{}}{ {44.2}\,($\pm$1.0) }\end{tabular}
& \begin{tabular}[c]{@{}c@{}}{ {63.0}\,($\pm$4.6) }\end{tabular}
& \begin{tabular}[c]{@{}c@{}}{ {33.2}\,($\pm$0.5) }\end{tabular}
& \begin{tabular}[c]{@{}c@{}}{ {61.6}\,($\pm$2.1) }\end{tabular}
& \begin{tabular}[c]{@{}c@{}}{ {37.8}\,($\pm$0.7) }\end{tabular}
& \begin{tabular}[c]{@{}c@{}}{ {60.3}\,($\pm$3.2) }\end{tabular}
\\
\multirow{1}{*}[0.2em]{\textbf{~~$+$\algname{}}}   
& \begin{tabular}[c]{@{}c@{}}{ \textbf{25.1}\,($\pm$0.3) }\end{tabular}
& \begin{tabular}[c]{@{}c@{}}{ \textbf{31.1}\,($\pm$1.2) }\end{tabular}
& \begin{tabular}[c]{@{}c@{}}{ \textbf{62.6}\,($\pm$0.7) }\end{tabular}
& \begin{tabular}[c]{@{}c@{}}{ \textbf{23.7}\,($\pm$1.0) }\end{tabular}
& \begin{tabular}[c]{@{}c@{}}{ \textbf{45.4}\,($\pm$0.5) }\end{tabular}
& \begin{tabular}[c]{@{}c@{}}{ \textbf{60.7}\,($\pm$5.1) }\end{tabular}
& \begin{tabular}[c]{@{}c@{}}{ \textbf{34.5}\,($\pm$0.7) }\end{tabular}
& \begin{tabular}[c]{@{}c@{}}{ \textbf{54.9}\,($\pm$3.2) }\end{tabular}
& \begin{tabular}[c]{@{}c@{}}{ \textbf{39.7}\,($\pm$0.4) }\end{tabular}
& \begin{tabular}[c]{@{}c@{}}{ \textbf{58.9}\,($\pm$4.1) }\end{tabular}
\\ \addlinespace[0.3ex]\cdashline{1-11}\addlinespace[0.5ex]
\multirow{1}{*}[0.2em]{~~Rel. Improv.}  
& \begin{tabular}[c]{@{}c@{}}{ 6.4\% }\end{tabular}
& \begin{tabular}[c]{@{}c@{}}{ 6.9\% }\end{tabular}
& \begin{tabular}[c]{@{}c@{}}{ 4.7\% }\end{tabular}
& \begin{tabular}[c]{@{}c@{}}{ 16.0\% }\end{tabular}
& \begin{tabular}[c]{@{}c@{}}{ 2.8\% }\end{tabular}
& \begin{tabular}[c]{@{}c@{}}{ 3.6\% }\end{tabular}
& \begin{tabular}[c]{@{}c@{}}{ 3.7\% }\end{tabular}
& \begin{tabular}[c]{@{}c@{}}{ 11.0\% }\end{tabular}
& \begin{tabular}[c]{@{}c@{}}{ 4.9\% }\end{tabular}
& \begin{tabular}[c]{@{}c@{}}{ 2.4\% }\end{tabular}
\\
\midrule

\multirow{1}{*}[0.2em]{MIR}                              

& \begin{tabular}[c]{@{}c@{}}{ {20.8}\,($\pm$0.6) }\end{tabular}
& \begin{tabular}[c]{@{}c@{}}{ {25.3}\,($\pm$1.7) }\end{tabular}
& \begin{tabular}[c]{@{}c@{}}{ {51.5}\,($\pm$0.5) }\end{tabular}
& \begin{tabular}[c]{@{}c@{}}{ {58.4}\,($\pm$1.3) }\end{tabular}
& \begin{tabular}[c]{@{}c@{}}{ {39.3}\,($\pm$1.2) }\end{tabular}
& \begin{tabular}[c]{@{}c@{}}{ {41.1}\,($\pm$2.0) }\end{tabular}
& \begin{tabular}[c]{@{}c@{}}{ {32.1}\,($\pm$0.4) }\end{tabular}
& \begin{tabular}[c]{@{}c@{}}{ {38.9}\,($\pm$3.1) }\end{tabular}
& \begin{tabular}[c]{@{}c@{}}{ {34.0}\,($\pm$1.0) }\end{tabular}
& \begin{tabular}[c]{@{}c@{}}{ {40.7}\,($\pm$1.6) }\end{tabular}
\\
\multirow{1}{*}[0.2em]{\textbf{~~$+$\algname{}}}
& \begin{tabular}[c]{@{}c@{}}{ \textbf{22.4}\,($\pm$0.1) }\end{tabular}
& \begin{tabular}[c]{@{}c@{}}{ \textbf{24.8}\,($\pm$1.0) }\end{tabular}
& \begin{tabular}[c]{@{}c@{}}{ \textbf{51.6}\,($\pm$0.6) }\end{tabular}
& \begin{tabular}[c]{@{}c@{}}{ \textbf{56.3}\,($\pm$1.6) }\end{tabular}
& \begin{tabular}[c]{@{}c@{}}{ \textbf{42.0}\,($\pm$0.9) }\end{tabular}
& \begin{tabular}[c]{@{}c@{}}{ \textbf{39.3}\,($\pm$4.5) }\end{tabular}
& \begin{tabular}[c]{@{}c@{}}{ \textbf{35.4}\,($\pm$1.4) }\end{tabular}
& \begin{tabular}[c]{@{}c@{}}{ \textbf{36.0}\,($\pm$3.1) }\end{tabular}
& \begin{tabular}[c]{@{}c@{}}{ \textbf{36.1}\,($\pm$0.6) }\end{tabular}
& \begin{tabular}[c]{@{}c@{}}{ \textbf{38.3}\,($\pm$3.4) }\end{tabular}

\\ \addlinespace[0.3ex]\cdashline{1-11}\addlinespace[0.5ex]
\multirow{1}{*}[0.2em]{~~Rel. Improv.}  
& \begin{tabular}[c]{@{}c@{}}{ 8.1\% }\end{tabular}
& \begin{tabular}[c]{@{}c@{}}{ 1.9\% }\end{tabular}
& \begin{tabular}[c]{@{}c@{}}{ 0.2\% }\end{tabular}
& \begin{tabular}[c]{@{}c@{}}{ 3.7\% }\end{tabular}
& \begin{tabular}[c]{@{}c@{}}{ 6.8\% }\end{tabular}
& \begin{tabular}[c]{@{}c@{}}{ 4.4\% }\end{tabular}
& \begin{tabular}[c]{@{}c@{}}{ 10.4\% }\end{tabular}
& \begin{tabular}[c]{@{}c@{}}{ 7.5\% }\end{tabular}
& \begin{tabular}[c]{@{}c@{}}{ 6.0\% }\end{tabular}
& \begin{tabular}[c]{@{}c@{}}{ 6.0\% }\end{tabular}
\\
\midrule

\multirow{1}{*}[0.2em]{GSS}                                               
& \begin{tabular}[c]{@{}c@{}}{ {21.8}\,($\pm$0.3) }\end{tabular}
& \begin{tabular}[c]{@{}c@{}}{ \textbf{27.9}\,($\pm$0.8) }\end{tabular}
& \begin{tabular}[c]{@{}c@{}}{ {46.0}\,($\pm$0.4) }\end{tabular}
& \begin{tabular}[c]{@{}c@{}}{ {69.7}\,($\pm$0.7) }\end{tabular}
& \begin{tabular}[c]{@{}c@{}}{ {40.3}\,($\pm$0.6) }\end{tabular}
& \begin{tabular}[c]{@{}c@{}}{ {71.5}\,($\pm$2.0) }\end{tabular}
& \begin{tabular}[c]{@{}c@{}}{ {34.5}\,($\pm$0.9) }\end{tabular}
& \begin{tabular}[c]{@{}c@{}}{ {65.6}\,($\pm$0.7) }\end{tabular}
& \begin{tabular}[c]{@{}c@{}}{ {35.3}\,($\pm$0.7) }\end{tabular}
& \begin{tabular}[c]{@{}c@{}}{ {64.2}\,($\pm$1.9) }\end{tabular}

\\
\multirow{1}{*}[0.2em]{\textbf{~~$+$\algname{}}}
& \begin{tabular}[c]{@{}c@{}}{ \textbf{23.7}\,($\pm$0.2) }\end{tabular}
& \begin{tabular}[c]{@{}c@{}}{ {28.4}\,($\pm$1.0) }\end{tabular}
& \begin{tabular}[c]{@{}c@{}}{ \textbf{49.1}\,($\pm$0.5) }\end{tabular}
& \begin{tabular}[c]{@{}c@{}}{ \textbf{69.2}\,($\pm$1.5) }\end{tabular}
& \begin{tabular}[c]{@{}c@{}}{ \textbf{41.3}\,($\pm$0.6) }\end{tabular}
& \begin{tabular}[c]{@{}c@{}}{ \textbf{68.5}\,($\pm$1.9) }\end{tabular}
& \begin{tabular}[c]{@{}c@{}}{ \textbf{36.5}\,($\pm$0.7) }\end{tabular}
& \begin{tabular}[c]{@{}c@{}}{ \textbf{61.6}\,($\pm$1.8) }\end{tabular}
& \begin{tabular}[c]{@{}c@{}}{ \textbf{35.9}\,($\pm$0.4) }\end{tabular}
& \begin{tabular}[c]{@{}c@{}}{ \textbf{62.3}\,($\pm$2.3) }\end{tabular}
\\ \addlinespace[0.3ex]\cdashline{1-11}\addlinespace[0.5ex]
\multirow{1}{*}[0.2em]{~~Rel. Improv.}  
& \begin{tabular}[c]{@{}c@{}}{ 8.8\% }\end{tabular}
& \begin{tabular}[c]{@{}c@{}}{ -1.7\% }\end{tabular}
& \begin{tabular}[c]{@{}c@{}}{ 6.5\% }\end{tabular}
& \begin{tabular}[c]{@{}c@{}}{ 0.6\% }\end{tabular}
& \begin{tabular}[c]{@{}c@{}}{ 2.3\% }\end{tabular}
& \begin{tabular}[c]{@{}c@{}}{ 4.2\% }\end{tabular}
& \begin{tabular}[c]{@{}c@{}}{ 5.9\% }\end{tabular}
& \begin{tabular}[c]{@{}c@{}}{ 6.1\% }\end{tabular}
& \begin{tabular}[c]{@{}c@{}}{ 1.8\% }\end{tabular}
& \begin{tabular}[c]{@{}c@{}}{ 2.9\% }\end{tabular}
\\
\midrule

\multirow{1}{*}[0.2em]{ASER}                                               
& \begin{tabular}[c]{@{}c@{}}{ {32.7}\,($\pm$0.1) }\end{tabular}
& \begin{tabular}[c]{@{}c@{}}{ {41.8}\,($\pm$0.4) }\end{tabular}
& \begin{tabular}[c]{@{}c@{}}{ {50.3}\,($\pm$0.5) }\end{tabular}
& \begin{tabular}[c]{@{}c@{}}{ {57.8}\,($\pm$3.3) }\end{tabular}
& \begin{tabular}[c]{@{}c@{}}{ {39.7}\,($\pm$0.7) }\end{tabular}
& \begin{tabular}[c]{@{}c@{}}{ {62.6}\,($\pm$2.1) }\end{tabular}
& \begin{tabular}[c]{@{}c@{}}{ {30.2}\,($\pm$0.2) }\end{tabular}
& \begin{tabular}[c]{@{}c@{}}{ {60.2}\,($\pm$0.9) }\end{tabular}
& \begin{tabular}[c]{@{}c@{}}{ {33.6}\,($\pm$0.5) }\end{tabular}
& \begin{tabular}[c]{@{}c@{}}{ {56.8}\,($\pm$1.1) }\end{tabular}

\\
\multirow{1}{*}[0.2em]{\textbf{~~$+$\algname{}}}
& \begin{tabular}[c]{@{}c@{}}{ \textbf{33.4}\,($\pm$0.4) }\end{tabular}
& \begin{tabular}[c]{@{}c@{}}{ \textbf{39.5}\,($\pm$0.8) }\end{tabular}
& \begin{tabular}[c]{@{}c@{}}{ \textbf{51.3}\,($\pm$0.3) }\end{tabular}
& \begin{tabular}[c]{@{}c@{}}{ \textbf{57.5}\,($\pm$0.8) }\end{tabular}
& \begin{tabular}[c]{@{}c@{}}{ \textbf{41.6}\,($\pm$0.8) }\end{tabular}
& \begin{tabular}[c]{@{}c@{}}{ \textbf{57.5}\,($\pm$3.2) }\end{tabular}
& \begin{tabular}[c]{@{}c@{}}{ \textbf{31.2}\,($\pm$0.2) }\end{tabular}
& \begin{tabular}[c]{@{}c@{}}{ \textbf{49.6}\,($\pm$2.8) }\end{tabular}
& \begin{tabular}[c]{@{}c@{}}{ \textbf{35.5}\,($\pm$0.4) }\end{tabular}
& \begin{tabular}[c]{@{}c@{}}{ \textbf{52.6}\,($\pm$2.5) }\end{tabular}
\\ \addlinespace[0.3ex]\cdashline{1-11}\addlinespace[0.5ex]
\multirow{1}{*}[0.2em]{~~Rel. Improv.}  
& \begin{tabular}[c]{@{}c@{}}{ 2.3\% }\end{tabular}
& \begin{tabular}[c]{@{}c@{}}{ 5.3\% }\end{tabular}
& \begin{tabular}[c]{@{}c@{}}{ 2.0\% }\end{tabular}
& \begin{tabular}[c]{@{}c@{}}{ 0.4\% }\end{tabular}
& \begin{tabular}[c]{@{}c@{}}{ 4.8\% }\end{tabular}
& \begin{tabular}[c]{@{}c@{}}{ 8.2\% }\end{tabular}
& \begin{tabular}[c]{@{}c@{}}{ 3.4\% }\end{tabular}
& \begin{tabular}[c]{@{}c@{}}{ 17.7\% }\end{tabular}
& \begin{tabular}[c]{@{}c@{}}{ 5.7\% }\end{tabular}
& \begin{tabular}[c]{@{}c@{}}{ 7.3\% }\end{tabular}
\\
\bottomrule
\end{tabular}
}
\vspace{-0.3cm}
\caption{Average accuracy\,(higher is better) and forgetting\,(lower is better) on Split CIFAR-100, Split CIFAR-10, and Split ImageNet-9 for a biased setup and on Split ImageNet-OnlyFG and Split ImageNet-Stylized for an unbiased setup. The best values are marked in bold.}
\label{tbl:overall_comparison_biased_unbiased_setup}
\vspace{-0.5cm}
\end{table*}

\subsection{Improvements through Debiasing}
\label{sec:perform_comparison}

\subsubsection{Biased Setup}
Table \ref{tbl:overall_comparison_biased_unbiased_setup} summarizes the performance of five replay-based OCL methods \emph{with} and \emph{without} applying \algname{} under the biased and unbiased setups.
Overall, \algname{} consistently improves their performances with respect to $A_{avg}$ and $F_{last}$. Quantitatively,  $A_{avg}$ and $F_{last}$ are improved by 5.3\% and 6.5\%, respectively, on average across all baselines and datasets. That is, the debiasing by \algname{} is very effective in OCL regardless of the method and dataset. In addition, the high $A_{avg}$ and the low $F_{last}$ demonstrate that the proposed debiasing approach helps expedite the training convergence of OCL models (higher transferability) and retain the knowledge of the previous tasks well (lower forgetability).


In detail, the effect of our method is the most prominent for ER, increasing $A_{avg}$ by 7.0\% on average. In particular, \algname{} improves the performances of ER and DER++ on Split CIFAR-10 and Split CIFAR-10 by 10.2\% and 9.4\% in terms of $A_{avg}$ and by 5.7\% and 16.0\% in terms of $F_{last}$, respectively. In contrast, ASER is improved relatively less. Since ASER is designed for keeping the samples with high diversity in the memory, we conjecture that the adverse effect of shortcut features is naturally smaller than random sampling as in ER and DER++. 

\def\arraystretch{0.65}
\begin{table}[t]
\scriptsize
\centering
\resizebox{1.0\columnwidth}{!}{%
\begin{tabular}[c]
{@{}l|C{0.8cm}C{0.8cm}|C{0.8cm}C{0.8cm}|C{0.8cm}C{0.8cm}c@{}}
\toprule
\multicolumn{1}{c|}{\textbf{}\hspace{-0.4cm}} & \multicolumn{2}{c}{\textbf{Split ImageNet-9}} & \multicolumn{2}{c}{\textbf{ImageNet-OnlyFG}} & \multicolumn{2}{c}{\textbf{ImageNet-Stylized}} \\
\multicolumn{1}{c|}{\textbf{\!\!\!\!\!\!\!\!\!\!\!\!\!\!\!\!Method\!\!\!\!\!\!\!\!\!\!\!\!\!}}
& \multicolumn{1}{c}{\hspace{-0.07cm}\scalebox{.9}{\bm{$A_{avg}$}\,($\uparrow$)}} 
& \multicolumn{1}{c|}{\hspace{-0.07cm}\scalebox{.9}{\bm{$F_{last}$}\,($\downarrow$)}}
& \multicolumn{1}{c}{\hspace{-0.07cm}\scalebox{.9}{\bm{$A_{avg}$}\,($\uparrow$)}}
& \multicolumn{1}{c|}{\hspace{-0.07cm}\scalebox{.9}{\bm{$F_{last}$}\,($\downarrow$)}}
& \multicolumn{1}{c}{\hspace{-0.07cm}\scalebox{.9}{\bm{$A_{avg}$}\,($\uparrow$)}}
& \multicolumn{1}{c}{\hspace{-0.07cm}\scalebox{.9}{\bm{$F_{last}$}\,($\downarrow$)}}
\\
\addlinespace[0.30ex]\toprule
\begin{tabular}[c]{@{}c@{}}\textrm{L2P\!\!\!}\end{tabular}
& \begin{tabular}[c]{@{}c@{}}{\hspace{-0.17cm}{95.3}\,($\pm$0.7) }\end{tabular}
& \begin{tabular}[c]{@{}c@{}}{\hspace{-0.17cm}{12.6}\,($\pm$1.9) }\end{tabular}
& \begin{tabular}[c]{@{}c@{}}{\hspace{-0.17cm}{91.0}\,($\pm$0.7) }\end{tabular}
& \begin{tabular}[c]{@{}c@{}}{\hspace{-0.17cm}{20.6}\,($\pm$2.2) }\end{tabular}
& \begin{tabular}[c]{@{}c@{}}{\hspace{-0.17cm}{87.7}\,($\pm$0.5) }\end{tabular}
& \begin{tabular}[c]{@{}c@{}}{\hspace{-0.17cm}{25.8}\,($\pm$1.8) }\end{tabular}
\\
\addlinespace[0.5ex]
 \begin{tabular}[c]{@{}c@{}}\textrm{$+$\textbf{\algname{}}\!\!\!}\end{tabular}
& \begin{tabular}[c]{@{}c@{}}{\hspace{-0.17cm}\textbf{97.1}\,($\pm$0.4) }\end{tabular}
& \begin{tabular}[c]{@{}c@{}}{\hspace{-0.17cm}\textbf{5.6}\,($\pm$1.1) }\end{tabular}
& \begin{tabular}[c]{@{}c@{}}{\hspace{-0.17cm}\textbf{92.9}\,($\pm$0.4) }\end{tabular}
& \begin{tabular}[c]{@{}c@{}}{\hspace{-0.17cm}\textbf{12.7}\,($\pm$0.6) }\end{tabular}
& \begin{tabular}[c]{@{}c@{}}{\hspace{-0.17cm}\textbf{89.7}\,($\pm$0.8) }\end{tabular}
& \begin{tabular}[c]{@{}c@{}}{\hspace{-0.17cm}\textbf{18.3}\,($\pm$1.0) }\end{tabular}
\\
\addlinespace[0.5ex]\cdashline{1-7}\addlinespace[0.6ex]
 \begin{tabular}[c]{@{}c@{}}\textrm{{Rel. Improv.}\!\!\!\!\! }\end{tabular}
& \begin{tabular}[c]{@{}c@{}}{\hspace{-0.07cm}1.9\% }\end{tabular}
& \begin{tabular}[c]{@{}c@{}}{\hspace{-0.07cm}55.4\% }\end{tabular}
& \begin{tabular}[c]{@{}c@{}}{\hspace{-0.07cm}2.1\% }\end{tabular}
& \begin{tabular}[c]{@{}c@{}}{\hspace{-0.07cm}38.3\% }\end{tabular}
& \begin{tabular}[c]{@{}c@{}}{\hspace{-0.07cm}2.3\% }\end{tabular}
& \begin{tabular}[c]{@{}c@{}}{\hspace{-0.07cm}28.9\% }\end{tabular}
\\
\midrule
\begin{tabular}[c]{@{}c@{}}\textrm{DualPrompt\!\!\!}\end{tabular}
& \begin{tabular}[c]{@{}c@{}}{\hspace{-0.17cm}{96.1}\,($\pm$0.6) }\end{tabular}
& \begin{tabular}[c]{@{}c@{}}{\hspace{-0.17cm}{9.9}\,($\pm$1.4) }\end{tabular}
& \begin{tabular}[c]{@{}c@{}}{\hspace{-0.17cm}{90.8}\,($\pm$1.2) }\end{tabular}
& \begin{tabular}[c]{@{}c@{}}{\hspace{-0.17cm}{20.0}\,($\pm$2.0) }\end{tabular}
& \begin{tabular}[c]{@{}c@{}}{\hspace{-0.17cm}{86.3}\,($\pm$1.5) }\end{tabular}
& \begin{tabular}[c]{@{}c@{}}{\hspace{-0.17cm}{29.2}\,($\pm$2.4) }\end{tabular}
\\
\addlinespace[0.5ex]
 \begin{tabular}[c]{@{}c@{}}\textrm{$+$\textbf{\algname{}}\!\!\!}\end{tabular}
& \begin{tabular}[c]{@{}c@{}}{\hspace{-0.17cm}\textbf{97.5}\,($\pm$0.3) }\end{tabular}
& \begin{tabular}[c]{@{}c@{}}{\hspace{-0.17cm}\textbf{3.7}\,($\pm$0.8) }\end{tabular}
& \begin{tabular}[c]{@{}c@{}}{\hspace{-0.17cm}\textbf{93.5}\,($\pm$0.5) }\end{tabular}
& \begin{tabular}[c]{@{}c@{}}{\hspace{-0.17cm}\textbf{9.8}\,($\pm$0.5) }\end{tabular}
& \begin{tabular}[c]{@{}c@{}}{\hspace{-0.17cm}\textbf{89.9}\,($\pm$0.6) }\end{tabular}
& \begin{tabular}[c]{@{}c@{}}{\hspace{-0.17cm}\textbf{16.6}\,($\pm$0.8) }\end{tabular}
\\ 
\addlinespace[0.5ex]\cdashline{1-7}\addlinespace[0.6ex]
 \begin{tabular}[c]{@{}c@{}}\textrm{{Rel. Improv.}\!\!\!\!\! }\end{tabular}
& \begin{tabular}[c]{@{}c@{}}{\hspace{-0.07cm}1.5\% }\end{tabular}
& \begin{tabular}[c]{@{}c@{}}{\hspace{-0.07cm}63.2\% }\end{tabular}
& \begin{tabular}[c]{@{}c@{}}{\hspace{-0.07cm}2.9\% }\end{tabular}
& \begin{tabular}[c]{@{}c@{}}{\hspace{-0.07cm}51.0\% }\end{tabular}
& \begin{tabular}[c]{@{}c@{}}{\hspace{-0.07cm}4.2\% }\end{tabular}
& \begin{tabular}[c]{@{}c@{}}{\hspace{-0.07cm}43.1\% }\end{tabular}
\\
\bottomrule
\end{tabular}
}
\vspace*{-0.3cm}
\caption{Performance of DropTop on top of pretrained ViT-based CL algorithms, L2P and DualPrompt, on Split ImageNet-9, Split ImageNet-OnlyFG, and Split ImageNet-Stylized.}
\vspace*{-0.6cm}
\label{tbl:perf_dualprompt}
\end{table}

\subsubsection{Unbiased Setup}
The improvement of \algname{} becomes more noticeable in the unbiased setup than in the biased setup, because this unbiased setup does not contain the shortcut features in the test datasets. For example, in Table \ref{tbl:overall_comparison_biased_unbiased_setup}, the relative improvement for MIR on Split ImageNet-OnlyFG is $10.4\%$, while that on Split ImageNet-9 is $6.8\%$.
%
In addition, \algname{} is more effective in reducing the background bias than the local cue bias; $A_{avg}$ improves by $6.6\%$ on average for all methods in Split ImageNet-OnlyFG, which is larger than $4.7\%$ in Split ImageNet-Stylized.
Achieving high accuracy for the unbiased setup requires robustly generalizing to more intrinsic complex features beyond less generalizable shortcut features. Obviously, attentive debiasing coordinated by adaptive intensity shifting supports the requirement by suppressing the undesirable reliance on shortcut features.




\vspace{-0.1cm}
\subsection{Debiasing Pretrained ViT-based CL}
\label{sec:dualprompt-droptop}

We test \algname{}'s adaptability to a pretrained ViT since it has been used frequently in recent CL studies, by injecting DropTop into L2P and DualPrompt in an online setting. Table \ref{tbl:perf_dualprompt} shows the performance of L2P and DualPrompt \emph{with} and \emph{without} \algname{} on the Split ImageNet datasets. We note that both versions of L2P or DualPrompt are modified to perform experience replay\,\cite{rolnick2019experience} using the small replay buffer for a fair comparison. Overall, \algname{} consistently exhibits significant improvements in both forgetting and accuracy, where the accuracy is initially high due to pretraining. Quantitatively, $F_{last}$ and $A_{avg}$ are improved by 46.7\% and 2.5\%, respectively, on average across the datasets and algorithms. This result clearly demonstrates the universal need for mitigating the shortcut bias in different backbone networks as well as \algname{}'s excellent adaptability. Please see Appendix F for the result on more datasets.

\def\arraystretch{0.5}
\begin{table}[t]
\scriptsize
\centering
\begin{tabular}[c]
{@{}l||c|c|c|cc@{}}
\toprule
\multicolumn{1}{c||}{\textbf{}} & \multicolumn{1}{c|}{\textbf{}} & \multicolumn{2}{c|}{\textbf{No Drop Int. Shift}} & \multicolumn{1}{c}{} \\
\multicolumn{1}{c||}{\textbf{Method}} 
& \multicolumn{1}{c|}{\textbf{No MF}\hspace{-1cm}}
& \multicolumn{1}{c|}{\textbf{Rand Drop}}
& \multicolumn{1}{c|}{\textbf{Fixed Drop}}
& \multicolumn{1}{c}{\textbf{\algname{}}}

\\ \toprule




\multirow{1}{*}[0.6em]{ER\hspace{-0.15cm}}                                
& \begin{tabular}[c]{@{}c@{}}{43.8\hspace{-1cm}}\end{tabular}
& \begin{tabular}[c]{@{}c@{}}{43.9}\end{tabular}
& \begin{tabular}[c]{@{}c@{}}{44.1}\end{tabular}
& \begin{tabular}[c]{@{}c@{}}{\textbf{45.5}}\end{tabular}
\\

\multirow{1}{*}[0.6em]{DER++\hspace{-0.15cm}}                                                   
& \begin{tabular}[c]{@{}c@{}}{42.0\hspace{-1cm}}\end{tabular}
& \begin{tabular}[c]{@{}c@{}}{41.6}\end{tabular}
& \begin{tabular}[c]{@{}c@{}}{43.4}\end{tabular}
& \begin{tabular}[c]{@{}c@{}}{\textbf{44.4}}\end{tabular}
\\

\multirow{1}{*}[0.6em]{MIR\hspace{-0.15cm}}                                               
& \begin{tabular}[c]{@{}c@{}}{{38.0}\hspace{-1cm}}\end{tabular}
& \begin{tabular}[c]{@{}c@{}}{{37.8}}\end{tabular}
& \begin{tabular}[c]{@{}c@{}}{{\textbf{39.3}}}\end{tabular}
& \begin{tabular}[c]{@{}c@{}}{{38.7}}\end{tabular}
\\

\multirow{1}{*}[0.6em]{GSS\hspace{-0.15cm}}                                 
& \begin{tabular}[c]{@{}c@{}}{34.7\hspace{-1cm}}\end{tabular}
& \begin{tabular}[c]{@{}c@{}}{34.8}\end{tabular}
& \begin{tabular}[c]{@{}c@{}}{37.0}\end{tabular}
& \begin{tabular}[c]{@{}c@{}}{\textbf{38.0}}\end{tabular}
\\

\multirow{1}{*}[0.6em]{ASER\hspace{-0.15cm}}                                               
& \begin{tabular}[c]{@{}c@{}}{40.5\hspace{-1cm}}\end{tabular}
& \begin{tabular}[c]{@{}c@{}}{41.4}\end{tabular}
& \begin{tabular}[c]{@{}c@{}}{40.5}\end{tabular}
& \begin{tabular}[c]{@{}c@{}}{\textbf{42.1}}\end{tabular}
\\
\midrule

\multirow{1}{*}[0.6em]{\textbf{AVG}\,\textbf{(Degrad.)}\hspace{-0.15cm}}                                               
& \begin{tabular}[c]{@{}c@{}}{{39.8}\,(\textbf{-4.9\%})\hspace{-1cm}}\end{tabular}
& \begin{tabular}[c]{@{}c@{}}{{39.9}\,(\textbf{-4.5\%})}\end{tabular}
& \begin{tabular}[c]{@{}c@{}}{{40.9}\,(\textbf{-2.1\%})}\end{tabular}
& \begin{tabular}[c]{@{}c@{}}{\textbf{{41.7}}}\end{tabular}
\\
\bottomrule

\end{tabular}
\vspace*{-0.3cm}
\caption{Ablation study on multi-level feature fusion and drop intensity shifting with respect to the accuracy $A_{avg}$ averaged over the biased datasets: Split CIFAR-100, Split CIFAR-10, and Split ImageNet-9. The highest values are marked in bold. }
\label{tbl:ablation}
\vspace*{-0.65cm}
\end{table}

\begin{figure*}[t!]
    \centering
    \includegraphics[width=\textwidth]{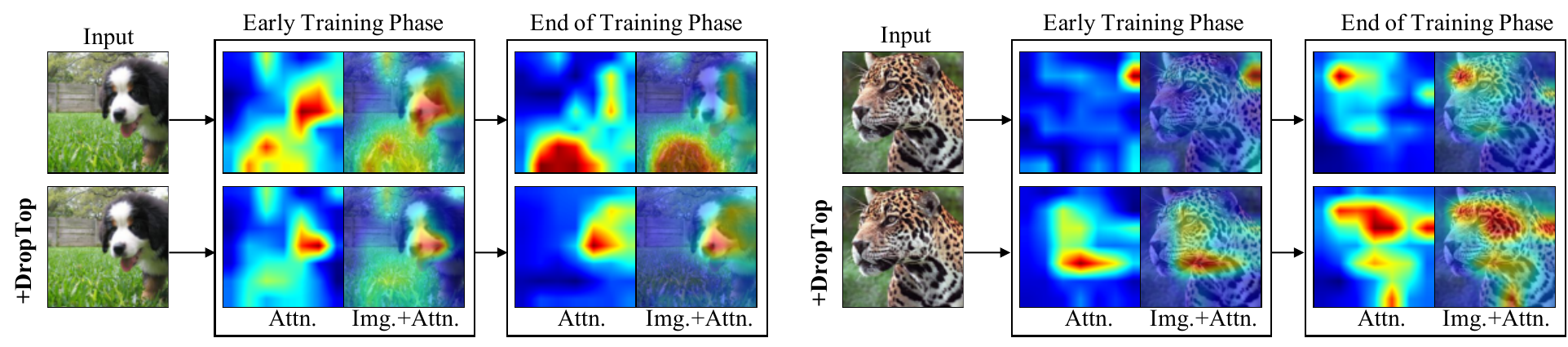} \vspace*{-0.6cm} \\
    {\small (a) Debiasing the \textbf{background} shortcut bias.} \hspace{3.6cm}  {\small (b) Debiasing the \textbf{local cue} bias.}
    \vspace*{-0.3cm}
    \caption{{Visualization of the gradual debiasing process by \algname{}}. The results are the activation maps from ER and ER$+$\algname{} trained on Split ImageNet-9: (a) and (b) are respectively related to debiasing the background and local cue bias during training, where the bias is the reliance on the grass background in (a) and on a local part of a leopard\,(e.g., part of the face without the body) in (b).} 
    \label{fig:activation_maps}
    \vspace*{-0.5cm}
\end{figure*}

\vspace{-0.1cm}
\subsection{Ablation Study}
\label{sec:ablation}
We conduct an ablation study to examine the general effectiveness of multi-level feature fusion in Eq.~\eqref{eq:feature_fusion} and adaptive intensity shifting in Eq.~\eqref{eq:shifting}. 
Table \ref{tbl:ablation} compares \algname{} with its three variants with respect to the average accuracy across the biased datasets (Split CIFAR-100, Split CIFAR-10, and Split ImageNet-9). The results for the unbiased datasets are presented in Appendix E. \emph{No MF}  drops only high-level features without multi-level fusion,  \emph{Rand Drop} drops randomly chosen features, and \emph{Fixed Drop} drops the features with the highest attention scores without adaptive intensity shifting.

\subsubsection{Multi-level Feature Fusion} The first variant, No MF, omits multi-level feature fusion for generating the drop masks, which solely relies on the high-level features. As a result, it shows the worst average accuracy among the variants. In particular, compared to \algname{} which relies on the first and last layers for multi-level feature fusion, the average accuracy drops significantly by 4.9\% on average. Therefore, refining semantic high-level features with structural low-level features via feature map fusion is essential to precisely identify the shortcut features. 

\subsubsection{Adaptive Intensity Shifting}
The other two variants, Rand Drop and Fixed Drop, do not apply adaptive intensity shifting while dropping the same proportion of the features as \algname{}. Based on their accuracy worse than \algname{}, we conclude that drop intensity shifting makes further improvements. Quantitatively, without drop intensity shifting, they face the average degradation of 4.5\% and 2.1\% compared with the complete \algname{}. 
Therefore, adaptively adjusting the drop intensity is needed for effective debiasing to catch the continuously varying proportion of the shortcut features in a timely manner.
Furthermore, the superior performance of \algname{} as well as Fixed Drop over Rand Drop verifies that debiasing based on highly activated features cannot be easily replaced by the regularization effect of the random drop\,\cite{srivastava2014dropout}.

\subsection{Qualitative Analysis for Debiasing Effect}
\label{sec:qualitative_result}

Figure \ref{fig:activation_maps} visualizes the activation maps of ER \emph{with} and \emph{without} \algname{} on Split ImageNet-9 to qualitatively analyze the debiasing effect. In summary, we observe that \algname{} gradually alleviates the shortcut bias as the training proceeds whereas the original ER increasingly relies on the shortcut bias---the background and local cue.
%
In Figure \ref{fig:activation_maps}(a) for the background bias, as the training proceeds, ER$+$\algname{} successfully debiases the grass background and, eventually, focuses on the intrinsic shapes of the dog at the end of training whereas ER rather intensifies the reliance on the background.
%
In Figure \ref{fig:activation_maps}(b) for the local cue bias, ER$+$\algname{} gradually expands the extent of the discriminative regions to cover the overall shapes of the object, resulting in more accurate and comprehensive recognition of the leopard compared with ER.
Please see Appendix H for more visualizations.

\begin{figure}[t!]
\centering
\includegraphics[width=1.05\columnwidth]{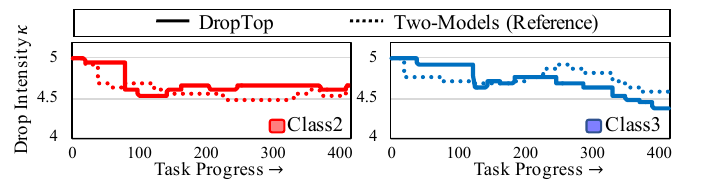}
\vspace*{-0.7cm}
\caption{Comparison between the original and two-model versions of \algname{} regarding the adjustments of the drop intensity $\kappa$ for Split CIFAR-10 on top of ER.}
\label{fig:intensity_hist}
\vspace*{-0.6cm}
\end{figure}


\subsection{Validity of Drop Intensity Adjustments}
\label{sec:correctness_intensity_shifting}

One might think that alternating the increment and decrement of $\kappa$ with a single model is invalid because the effect of one option may influence that of the other option. Using only a single model is inevitable because of the memory restriction in OCL, as discussed in Section \ref{subsec:AdaptiveIntensityShifting}. Thus, by assuring the period for each option to be long enough, we aim to reducing undesirable influence between the two options. To verify the validity of our approach, we implement a two-model ``reference'' version that maintains two separate models, each for increment and decrement, but does violate the memory restriction. As shown in Figure \ref{fig:intensity_hist}, the original version adjusts $\kappa$ very similarly to the reference version. Quantitatively, {82.1}{\tiny$\pm$0.6}\% of the adjustments of $\kappa$ coincide with each other, and their accuracy is nearly identical, {61.0}{\tiny$\pm$0.6}\% and {61.5}{\tiny$\pm$0.2}\% for the test. 
\section{Conclusion}
\label{sec:conclusion}

We propose a debiasing OCL method called \textbf{\algname{}}, introducing two novel solutions for debiasing shortcut features, \emph{attentive debiasing} with \emph{feature map fusion} and \emph{adaptive intensity shifting}. Without relying on prior knowledge and auxiliary data, \algname{} determines the appropriate level and proportion of possible short features and drops them from the feature map for debiasing. It can easily be built on top of any existing replay-based OCL methods. 
The evaluation confirms that \algname{} significantly improves the existing state-of-the-art OCL methods. Overall, we believe that our work sheds the light on the importance of debiasing shortcuts in OCL.


\section*{Acknowledgements}
This work was supported by Institute of Information \& Communications Technology Planning \& Evaluation\,(IITP) grant funded by the Korea government\,(MSIT) (No.\ 2022-0-00157, Robust, Fair, Extensible Data-Centric Continual Learning).

\bibliography{aaai24}

\clearpage

\begin{center}
  \LARGE 
  Adaptive Shortcut Debiasing for \\ Online Continual Learning \par
  \large 
  Appendix
  
  \vspace{1.0em}
  
  
  
\end{center}

\section{A \quad Pseudocode of \algname{}}

\algsetup{linenosize=\small}\newlength{\oldtextfloatsep}\setlength{\oldtextfloatsep}{\textfloatsep}
\setlength{\textfloatsep}{10pt}
\begin{algorithm}[b!]
\small
\caption{\small Replay-based OCL with \algname{}}
\label{alg:pseudocode}
\begin{algorithmic}[1]
\REQUIRE {Number of tasks $T$, online data stream $\{D_1,\dots,D_T\}$, total drop ratio $\gamma$, softmax function $\sigma$, learning rate $\eta$, batch size $b$
}
\ENSURE {Model parameters $\theta, w$ after learning task $T$}
\STATE {$\theta, w \leftarrow \text{Initialize model parameters};$}
\label{lst:line:init_s}
\STATE {$\mathcal{B} \leftarrow \text{Initialize replay memory};$}
\STATE {$\kappa \leftarrow \text{Initialize class-wise debiasing intensity};$}
\label{lst:line:init_e}
\FOR {$t=1$ {\bf to} $T$}
\FOR {$i=1$ {\bf to} $|D_t|/b$}
\STATE {${X}_n, {Y}_n$ $\leftarrow$ a mini-batch from online data stream $D_t$;} 
\label{lst:line:online_batch}
\STATE {\COMMENT{\textsc{Replay Memory Retrieval}}}
\STATE {${X}_m, {Y}_m$ $\leftarrow$ ${\rm MemoryRetrieve}(\mathcal{B}, {X}_n, {Y}_n)$;} 
\label{lst:line:memory_batch}
\STATE {\COMMENT{\textbf{\textsc{Attentive Debiasing in Sec. 4.1}}}}
\label{lst:line:online_debiasing_layer}
\STATE {$M$ $\leftarrow$ ${\rm MultiLevelDropMask}({X}_n\cup{X}_m;\theta,\kappa,\gamma)$;}\hspace*{-1cm}
\label{lst:line:drop_mask}
\STATE {$P$ $\leftarrow$ $\sigma({\rm DebiasForward}({X}_n\cup{X}_m, M; \theta, w)) $;} 
\label{lst:line:debias_forward}
\STATE {\COMMENT{\textsc{Model Update}}}
\label{lst:line:update_s}
\STATE {Compute the loss $\mathcal{J}$ using ${Y}_n\cup{Y}_m$ and ${P}$;}
\STATE {$\theta \leftarrow \theta - \eta\nabla_{\theta}\mathcal{J}$;~~$w \leftarrow w - \eta\nabla_{w}\mathcal{J}$;}
\label{lst:line:update_e}
\STATE {\COMMENT{\textsc{Replay Memory Update}}}
\label{lst:line:mem_update_s}
\STATE {$\mathcal{B}$ $\leftarrow$ ${\rm MemoryUpdate}(\mathcal{B}, {X}_n, {Y}_n)$;} 
\label{lst:line:mem_update_e}
\STATE {\COMMENT{\textbf{\textsc{Adaptive Intensity Shifting in Sec. 4.2}}}}
\label{lst:line:adjust_s}
\STATE {$\kappa$ $\leftarrow$ ${\rm AdjustDropIntensity}(\mathcal{M}, i)$;} 
\label{lst:line:adjust_e}
\ENDFOR
\ENDFOR
\RETURN {$\theta,w$;}
\end{algorithmic}
\end{algorithm}
Algorithm \ref{alg:pseudocode} describes how \algname{} works with replay-based online continual learning\,(OCL) methods. Various replay-based algorithms are applied by using their respective methods for \textit{MemoryRetrieve}\,(Line \ref{lst:line:memory_batch}) and \textit{MemoryUpdate}\,(Line \ref{lst:line:mem_update_s}).
At the beginning, a model and a memory buffer are initialized for a replay-based algorithm; and class-wise drop intensity is initialized for our \algname{}\,(Lines \ref{lst:line:init_s}--\ref{lst:line:init_e}). The model receives two minibatches from the online data stream and the memory buffer, respectively\,(Lines \ref{lst:line:online_batch}--\ref{lst:line:memory_batch}). 
The forward propagation is carried out through the online debiasing layer\,(Lines \ref{lst:line:online_debiasing_layer}--\ref{lst:line:debias_forward}). First, a drop mask is obtained by masking top $\kappa$\% and random $(\gamma-\kappa)$\% of the multi-level features\,(Line \ref{lst:line:drop_mask}). Next, the drop mask is applied to the forward pass by multiplying it element-by-element with the intermediate feature maps\,(Line \ref{lst:line:debias_forward}).
Then, based on the cross-entropy loss with the combined batch, the backward propagation is carried out to update the model parameters\,(Lines \ref{lst:line:update_s}--\ref{lst:line:update_e}).
In addition, the replay algorithm performs an online update of the memory buffer using the minibatch from the data stream\,(Lines \ref{lst:line:mem_update_s}--\ref{lst:line:mem_update_e}).
Last, the class-wise drop intensity for each class is updated\,(Lines \ref{lst:line:adjust_s}--\ref{lst:line:adjust_e}). The detailed pseudocode of adaptive intensity shifting is described in Algorithm \ref{alg:pseudocode_adaptive_scheduler}, which is self-explanatory.


\algsetup{linenosize=\small}
\setlength{\textfloatsep}{8pt}
\begin{algorithm}[!t]
\small
\caption{\small Adaptive Intensity Shifting}
\label{alg:pseudocode_adaptive_scheduler}
\begin{algorithmic}[1]
\footnotesize 
\REQUIRE {Preceding intensity: $\kappa^\prime$, step size: $\alpha$, alternating period: $p$, episodic memory of each class: $\mathcal{B}$, model parameters: $\{\theta,w\}$, training iteration: $i$
}
\ENSURE {Drop intensity $\kappa$ that will be applied to the next model update}
\STATE {\COMMENT{{\textsc{Initialize candidate strategies}}}}
\STATE {$\kappa^{dec} \leftarrow \kappa^\prime*\alpha; ~~\kappa^{inc} \leftarrow \kappa^\prime * (1/\alpha);$}
\STATE {\{$\mathcal{H}^{dec}, \mathcal{H}^{inc}\} \leftarrow {\rm InitializeHistory}()$;}
\IF{${\rm mod}(i,p)=0$} 
\STATE {\COMMENT{{\textsc{Compute contribution of intensity}}}}
\STATE {$\mathcal{L}_{new}$ $\leftarrow$ ${\rm Loss}(\mathcal{B}, \{\theta, w\})$;}
\STATE {$\Delta\mathcal{L}\leftarrow \mathcal{L}_{old}-\mathcal{L}_{new}$;} 
\STATE { ${\rm UpdateHistory}(\{\mathcal{H}^{dec}, \mathcal{H}^{inc}\}, \Delta\mathcal{L})$;} 
\STATE {$\mathcal{L}_{old}$ $\leftarrow$ $\mathcal{L}_{new}$;} 
\STATE {\COMMENT{{\textsc{Update Intensity based on Statistical Test}}}}
\IF {${\rm full}(\mathcal{H}^{dec}, \mathcal{H}^{inc})$}
\IF {${\rm \textit{p}\text{-}value}(\mathcal{H}^{dec}, \mathcal{H}^{inc})\le0.05$}
    \STATE {$\kappa^{inc}$ $\leftarrow$ $\kappa^{dec}$;}
    \STATE {$\kappa^{dec}$ $\leftarrow$ $\kappa^{dec}*\alpha$;}
\ELSIF{${\rm \textit{p}\text{-}value}(\mathcal{H}^{dec}, \mathcal{H}^{inc})\ge0.95$}
    \STATE {$\kappa^{dec}$ $\leftarrow$ $\kappa^{inc}$;}
    \STATE {$\kappa^{inc}$ $\leftarrow$ $\kappa^{inc}*(1/\alpha)$;}
\ENDIF
\STATE {${\rm ResetHistory}(\mathcal{H}^{dec}, \mathcal{H}^{inc})$;}
\ENDIF
\STATE {\COMMENT{{\textsc{Alternate two shift directions} }}}
\STATE {$\kappa \leftarrow {\rm AlternateBetween}(\kappa^{dec}, \kappa^{inc},i)$ ;} 
\ELSE 
\STATE {$\kappa \leftarrow \kappa^\prime$;} 
\ENDIF
\RETURN {$\kappa$;}
\end{algorithmic}
\end{algorithm}

\section{B \quad Empirical Proofs of Low Transferability and High Forgetting}
\label{sec:empirical_shorcut_bias_ocl}

\begin{figure}[!t]
    \centering
    \includegraphics[width=0.9\columnwidth]{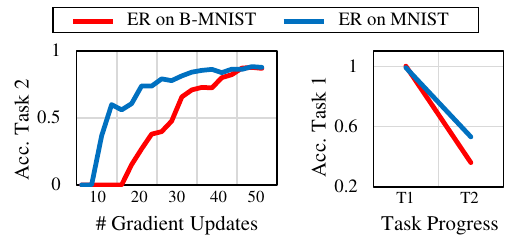} \vspace*{-0.05cm} \\
    \small \hspace*{0.9cm} (a) Transferability. \hspace{1.2cm} (b) Forgetability.
    \vspace{-0.2cm}
    \caption{Shortcut learning in OCL: (a) shows lower transferability from Task 1\,({\color{red}0} and {\color{green}1}) to Task 2\,({\color{green}2} and {\color{green}3}), and (b) shows high forgetting of Task 1\,({\color{green}0} and {\color{green}1}) by Task 2\,({\color{red}2} and {\color{green}3}), compared with non-shortcut learning.}
    \vspace*{-0.1cm}
    \label{fig:preliminary_exp}
\end{figure}

To empirically investigate transferability and forgetting, we conduct a simple controlled experiment using Biased MNIST (B-MNIST)\,\cite{bahng2020learning}, which is a widely-used dataset in bias learning\,\cite{bahng2020learning,bui2021exploiting,tartaglione2021end}, as well as  MNIST\,\cite{lecun1998gradient}.
In B-MNIST, each digit is in a different color, e.g., 0 in red and 1 in green.
DNNs trained on B-MNIST employ only color shortcut features for classification because the color is a simpler feature than the shape of a digit. We train ER\,\cite{rolnick2019experience} for OCL with two tasks---Task\,1 (T1) classifying 0 and 1, and Task 2\,(T2) classifying 2 and 3.

\smallskip\noindent{\textbf{Low Transferability.}}
Figure \ref{fig:preliminary_exp}(a) shows that shortcut features have lower transferability than non-shortcut features. When training T2, ER trained on B-MNIST (the shortcut-biased model) shows slower increase of test accuracy than ER trained on MNIST, since the red and green color features learned in T1 are not applicable to T2.
This result indicates that the transferability of the OCL model significantly deteriorates once the model is biased towards the shortcut features.

\smallskip\noindent{\textbf{High Forgetability.}}
Figure \ref{fig:preliminary_exp}(b) shows that shortcut features have higher forgetting than non-shortcut features.
After training T2, ER trained on B-MNIST shows rapid forgetting of the prediction capability in T1, since the red or green color becomes a shortcut cue to predict the digit 2 or 3 in T2.
This result indicates that the OCL model is also significantly damaged by forgetting once the model is biased towards the shortcut features.


\def\arraystretch{0.4}
\begin{table}[!t]
\small
\centering
\resizebox{1.0\columnwidth}{!}{%
\begin{tabular}[c]
{@{}c|cccc|cc|cc@{}}
\toprule
\multicolumn{1}{c|}{\textbf{}\hspace{-0.4cm}} & \multicolumn{6}{c|}{\textbf{{OCL}}} & \multicolumn{1}{c}{\textbf{{Standard}}} \\
\cline{2-8}\addlinespace[0.20ex]
\multicolumn{1}{c|}{\textbf{\!\!\!\!\!\!\!\!\!\!\!\!\!\!\!\!Dataset\!\!\!\!\!\!\!\!\!\!\!\!\!}}
& \multicolumn{1}{c}{\hspace{-0.07cm}\scalebox{.7}{$Task_1$}} 
& \multicolumn{1}{c}{\hspace{-0.07cm}\scalebox{.7}{$Task_2$}}
& \multicolumn{1}{c}{\hspace{-0.07cm}\scalebox{.7}{$Task_3$}}
& \multicolumn{1}{c|}{\hspace{-0.07cm}\scalebox{.7}{$Task_4$}}
& \multicolumn{1}{c}{\hspace{-0.07cm}\scalebox{.7}{\bm{$A_{avg}$}\,($\uparrow$)\!\!}}
& \multicolumn{1}{c|}{\hspace{-0.07cm}\scalebox{.7}{\bm{$F_{last}$}\,($\downarrow$)\!\!}}
& \multicolumn{1}{c}{\hspace{-0.07cm}\scalebox{.7}{\bm{$Acc.$}\,($\uparrow$)}}
\\
\addlinespace[0.30ex]\toprule
\begin{tabular}[c]{@{}c@{}}{\textbf{OnlyBG}\!}\end{tabular}
& \begin{tabular}[c]{@{}c@{}}{\hspace{-0.17cm}\textbf{54.4}\scalebox{.75}{$\pm$2.2} }\end{tabular}
& \begin{tabular}[c]{@{}c@{}}{\hspace{-0.17cm}{35.4}\scalebox{.75}{$\pm$1.7} }\end{tabular}
& \begin{tabular}[c]{@{}c@{}}{\hspace{-0.17cm}{21.6}\scalebox{.75}{$\pm$1.6} }\end{tabular}
& \begin{tabular}[c]{@{}c@{}}{\hspace{-0.17cm}{16.3}\scalebox{.75}{$\pm$0.8}\!\!\!\! }\end{tabular}
& \begin{tabular}[c]{@{}c@{}}{\hspace{-0.17cm}{31.9}\scalebox{.75}{$\pm$1.2}\!\! }\end{tabular}
& \begin{tabular}[c]{@{}c@{}}{\hspace{-0.4cm}{45.6}\scalebox{.75}{$\pm$1.7}\!\!\!\!\!\!\! }\end{tabular}
& \begin{tabular}[c]{@{}c@{}}{\hspace{-0.17cm}{28.4}\scalebox{.75}{$\pm$1.6} }\end{tabular}
\\
\addlinespace[0.5ex]
 \begin{tabular}[c]{@{}c@{}}{\textbf{OnlyFG}\!}\end{tabular}
& \begin{tabular}[c]{@{}c@{}}{\hspace{-0.17cm}{54.0}\scalebox{.75}{$\pm$0.3} }\end{tabular}
& \begin{tabular}[c]{@{}c@{}}{\hspace{-0.17cm}\textbf{37.3}\scalebox{.75}{$\pm$2.3} }\end{tabular}
& \begin{tabular}[c]{@{}c@{}}{\hspace{-0.17cm}\textbf{24.9}\scalebox{.75}{$\pm$0.6} }\end{tabular}
& \begin{tabular}[c]{@{}c@{}}{\hspace{-0.17cm}\textbf{21.2}\scalebox{.75}{$\pm$0.6}\!\!\!\! }\end{tabular}
& \begin{tabular}[c]{@{}c@{}}{\hspace{-0.17cm}\textbf{34.4}\scalebox{.75}{$\pm$0.6}\!\! }\end{tabular}
& \begin{tabular}[c]{@{}c@{}}{\hspace{-0.4cm}\textbf{40.5}\scalebox{.75}{$\pm$1.6}\!\!\!\!\!\!\! }\end{tabular}
& \begin{tabular}[c]{@{}c@{}}{\hspace{-0.17cm}\textbf{29.1}\scalebox{.75}{$\pm$0.4} }\end{tabular}
\\
\addlinespace[0.5ex]\cdashline{1-8}\addlinespace[0.6ex]
 \begin{tabular}[c]{@{}c@{}}\textrm{{Rel. Improv.}\!\!\! }\end{tabular}
& \begin{tabular}[c]{@{}c@{}}{\hspace{-0.07cm}-0.8\% }\end{tabular}
& \begin{tabular}[c]{@{}c@{}}{\hspace{-0.07cm}5.5\% }\end{tabular}
& \begin{tabular}[c]{@{}c@{}}{\hspace{-0.07cm}15.4\% }\end{tabular}
& \begin{tabular}[c]{@{}c@{}}{\hspace{-0.07cm}29.6\% }\end{tabular}
& \begin{tabular}[c]{@{}c@{}}{\hspace{-0.07cm}7.6\% }\end{tabular}
& \begin{tabular}[c]{@{}c@{}}{\hspace{-0.07cm}11.1\% }\end{tabular}
& \begin{tabular}[c]{@{}c@{}}{\hspace{-0.07cm}2.7\% }\end{tabular}
\\
\bottomrule
\end{tabular}
}
\vspace*{-0.2cm}
\caption{Impact of shortcut bias on OCL and standard learning.}
\label{tbl:shortcut_on_ocl}
\end{table}

\smallskip\noindent{\textbf{Impact of Realistic Shortcuts on OCL.}}
Table \ref{tbl:shortcut_on_ocl} compares the performance of DNNs in OCL and standard learning \emph{with and without} background shortcut features.
In longer OCL with real-world images, the shortcut bias on background significantly degrades the performance. Moreover, the negative impact of the shortcut bias is much more pronounced in OCL than in standard learning.
In detail, to enable or disable the shortcut bias, ER was trained using two ImageNet-variant datasets\,\cite{xiao2020noise}: OnlyBackground\,(OnlyBG) and OnlyForeground\,(OnlyFG).
ER trained on OnlyBG experienced a more severe degradation in terms of accuracy\,(transferability) and forgetting, by 7.6\% and 11.1\% respectively, compared with ER trained on OnlyFG. In standard learning, the accuracy gap between two datasets is much smaller at 2.7\%.

\section{C \quad Split ImageNet-9 Datasets}

ImageNet-9\,\cite{xiao2020noise}\footnote{The dataset is obtained at \url{https://github.com/MadryLab/backgrounds_challenge}.}, a subset of ImageNet\,\cite{deng2009imagenet}, contains 9 coarser-level classes: Dog, Bird, Vehicle, Reptile, Carnivore, Insect, Instrument, Primate, and Fish. Additionally, the ratio of the classes is balanced, and each class has about 4,500 instances.
See Appendix A of \cite{xiao2020noise} for the further details of ImageNet-9.

\smallskip\noindent{\textbf{Task Configuration.}}
\textit{Split ImageNet-9} is constructed for the continual learning environment by evenly distributing the 9 classes into 4 tasks except the class \textit{Primate}. Thus, the 4 sequential tasks are  (Dog, Bird), (Vehicle, Reptile), (Carnivore, Insect), and (Instrument, Fish). 

\smallskip\noindent{\textbf{Split ImageNet-9 Variants for the Unbiased Setup.}}
The two variants are used to measure the debiasing efficacy of \algname{}. 
\textit{Split ImageNet-OnlyFG} is created by adopting the same split of ImageNet as originally done in \cite{xiao2020noise}. Similarly, Split ImageNet-Stylized is created by applying the same split of Stylized ImageNet as done in \cite{geirhos2018imagenet}.

\section{D \quad Hyperparameter Sensitivity Analysis}

\begin{figure}[!t]
    \centering
    \includegraphics[width=0.98\columnwidth]{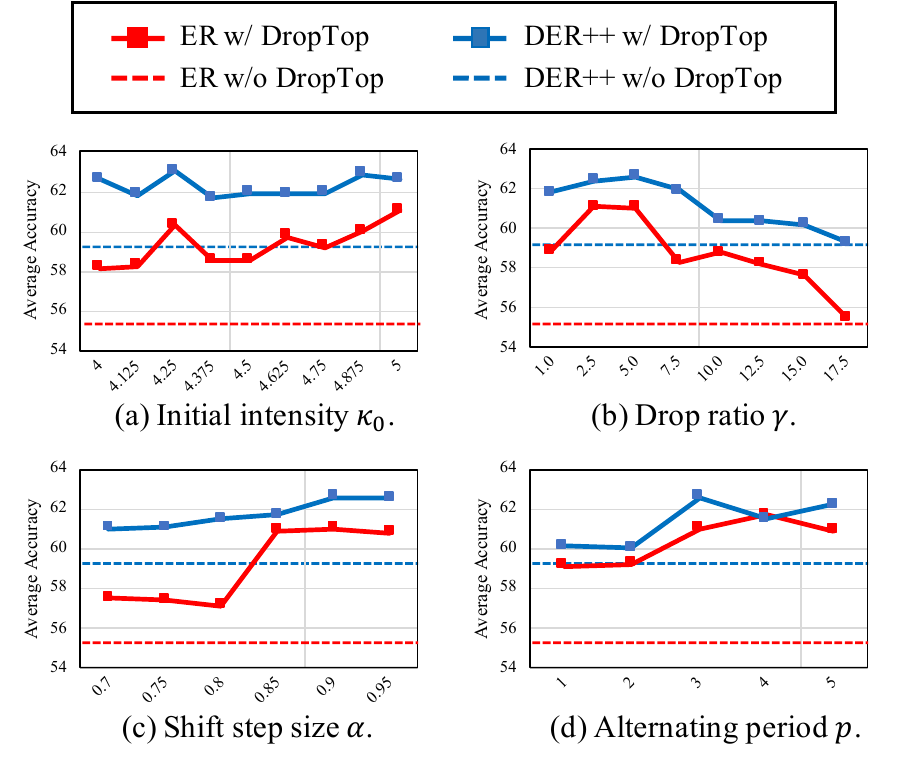}
    \vspace{-0.3cm}
    \caption{Parameter sensitivity analysis: 
    the default values are the initial drop intensity $\kappa_0=5.0$\%, the drop ratio $\gamma=5.0$\%, the shifting step size $\alpha=0.9$, and the alternating period $p=3$. Each subfigure shows how the average accuracy $A_{avg}$ of five runs changes as only one of the hyperparamters varies.
    }
    \label{fig:parameter_sensitivity}
\end{figure}

The hyperparameters of \algname{} are the initial drop intensity $\kappa_0$, the total drop ratio $\gamma$, the shifting step size $\alpha$, and the alternating period $p$. Figure \ref{fig:parameter_sensitivity} shows how each hyperparameter affects the average accuracy with ER\,\cite{rolnick2019experience} and DER\,\cite{buzzega2020dark}.  

\smallskip\noindent{\textbf{Initial Drop Intensity.}}
The initial drop intensity $\kappa_0$ is continuously shifted during the training period coordinated by adaptive intensity shifting. Figure \ref{fig:parameter_sensitivity}(a) shows that the performance is consistently high over a wide range of $\kappa_0$, and we conjecture that the continuous adjustment of the intensity leads to such consistency. Among similarly good values, we set $\kappa_0=5.0$\% at which ER and DER++ peak, for all experiments of these algorithms.
The values of the hyperparameters for the other algorithms are chosen in the same way.

\smallskip\noindent{\textbf{Drop Ratio.}}
A proper drop ratio $\gamma$ prevents too much removal of informative features and too small debiasing effect. Interestingly, Figure \ref{fig:parameter_sensitivity}(b) shows that the performance begins to drop if $\gamma$ is greater than $7.5$\% while losing the benefit from \algname{} if $\gamma$ becomes $17.5$\%; thus, we set $\gamma=5.0$\% as the proper drop ratio for all experiments.

\smallskip\noindent{\textbf{Shifting Step Size.}}
The shifting step size $\alpha$ determines the size of a single intensity update to handle the varying extent of the shortcut bias. The smaller $\alpha$, the larger a shifting step size.
Figure \ref{fig:parameter_sensitivity}(c) shows that \algname{} achieves consistently high performance as long as $\alpha$ is greater than 0.85, and particularly we set $\alpha=0.9$ where the performance peaks, for all experiments. For a small $\alpha$\,(a large shifting step size), the drop intensity may not be finely adjusted, resulting in suboptimal performances.

\smallskip\noindent{\textbf{Alternating Period.}}
The alternating period $p$ decides how often the contribution of each intensity to the loss reduction is measured. If two intensities alternate too often with a small $p$, measuring the contribution of the intensity will become inconsistent and inaccurate. In this sense, we observe consistently high performances when $p$ is greater than or equal to 3, as in Figure \ref{fig:parameter_sensitivity}(d). Therefore, $p$ is decided to 3 for all experiments. 

\section{E \quad Complete Experiment Results}
\subsection{Implementation Settings}
For all algorithms and datasets, the size of a minibatch from the data stream and the replay memory is set to 32\,\cite{buzzega2020dark}. The size of episodic memory is set to 500 for Split CIFAR-10 and Split ImageNet-9 and 2,000 for Split CIFAR-100 depending on the total number of classes. In order to add \algname{}'s debiasing capability to L2P and DualPrompt, which are originally rehearsal-free, we attach small replay memory whose size is reduced to 200 and 500, respectively. 
We train ResNet18 using SGD with a learning rate of 0.1\,\cite{buzzega2020dark, shim2021online} for all ResNet-based algorithms. We optimize L2P and DualPrompt with a pretrained ViT-B/16 using Adam with a learning rate of 0.05, $\beta_1$ of 0.9, and $\beta_2$ of 0.999.

Since we follow the widely-used OCL setting of \cite{shim2021online, buzzega2020dark} regarding the batch sizes, optimizer, backbone, etc., we adopt the hyperparameter values selected through an extensive grid search in the same OCL setting\,\cite{shim2021online, buzzega2020dark}. Specifically, we adopt the values of ER and DER++ following \cite{buzzega2020dark}\footnote{\url{https://github.com/aimagelab/mammoth}}, those of MIR, GSS, and ASER following \cite{shim2021online}\footnote{\url{https://github.com/RaptorMai/online-continual-learning}}, and those of L2P and DualPrompt following \cite{wang2022dualprompt}\footnote{\url{https://github.com/JH-LEE-KR/dualprompt-pytorch}}. 

The hyperparameters specific to the individual algorithms are set for all datasets as follows:
\begin{itemize}[leftmargin=10pt,noitemsep]
\item {ER}\,\cite{rolnick2019experience}: No additional hyperparameters.
\item {DER++}\,\cite{buzzega2020dark}: The coefficient of knowledge distillation loss is set to 0.2, and the coefficient of the cross entropy loss using a batch from the memory buffer is set to 0.5.
\item {MIR}\,\cite{mir}: The number of subsamples for searching maximally interfered samples is set to 50.
\item {GSS}\,\cite{aljundi2019gradient}: The number of batches from the memory buffer to estimate the maximal similarity score is set to 10.
\item {ASER}\,\cite{shim2021online}: The mean values of Adversarial SV and Cooperative SV are used as a type of ASER. The number of samples per class is set to 1.5, and the number of neighbors for computing KNN-SV is set to 3. 
\item {L2P}\,\cite{wang2022learning}: The prompt length, top-N, and size of the prompt pool are set to 5, 5, and 10, respectively. Besides, to provide an optimization similar to DualPrompt\,\cite{wang2022dualprompt}, we apply prefix tuning-based prompts to the first through fifth Transformer layers.
\item {DualPrompt}\,\cite{wang2022dualprompt}: The general prompts with a length of 5 are applied to the first and second Transformer layers, while the expert prompts with a length of 20 are applied to the third, fourth, and fifth Transformer layers. 
\end{itemize}
For the empirical proofs in Appendix B, we use ER with the ResNet18 backbone and SGD with an initial learning rate of 0.001. The size of episodic memory is set to 200.

\subsection{Hyperparameters}
There are a few hyperparameters introduced by \algname{}. For attentive debiasing, we fix the total drop ratio $\gamma$ to $5.0\%$ and set the initial drop intensity $\kappa_0$ to $5.0\%$ for ER, DER++, and MIR and to $0.5\%$ for GSS and ASER, differently depending on the sampling method. For L2P and DualPrompt, we set $\gamma$ and $\kappa_0$ to $2.0\%$ and $1.0\%$, respectively, owing to the difference of the backbone network. 
For adaptive intensity shifting, we fix the history length $l$ to $10$, which is approximately the minimum sample size for one-sided $t$-test with the significance level 0.05\,\cite{heckert2002ttest}. 
The alternating period $p = 3$ and the shifting step size $\alpha = 0.9$ are adequate across the algorithms and datasets.
Please see Appendix D for the sensitivity analysis of the hyperparameters and Appendix E for the hyperparameter tuning\,\cite{buzzega2020dark} for the other algorithms.

\def\arraystretch{0.8}
\begin{table}[t]
\scriptsize
\centering
\begin{tabular}[c]
{@{}l||cc|ccc@{}}
\toprule
\multicolumn{1}{c||}{\textbf{Method}} 
& \multicolumn{1}{c}{\textbf{\bm{$A_{avg}$}\,($\uparrow$)}}
& \multicolumn{1}{c|}{\textbf{Degrade}}
& \multicolumn{1}{c}{\bm{$F_{last}$}\,($\downarrow$)}
& \multicolumn{1}{c}{\textbf{Degrade}}
\\ \toprule
\multirow{1}{*}[0.3em]{{Soft Drop Mask}}                                
& \begin{tabular}[c]{@{}c@{}}{60.3\,($\pm$0.4)}\end{tabular}
& \begin{tabular}[c]{@{}c@{}}{1.1\%}\end{tabular}
& \begin{tabular}[c]{@{}c@{}}{40.4\,($\pm$3.1)}\end{tabular}
& \begin{tabular}[c]{@{}c@{}}{7.4\%}\end{tabular}
\\
\multirow{1}{*}[0.3em]{{Common Drop Intensity}}                                                   
& \begin{tabular}[c]{@{}c@{}}{57.6\,($\pm$0.1)}\end{tabular}
& \begin{tabular}[c]{@{}c@{}}{5.6\%}\end{tabular}
& \begin{tabular}[c]{@{}c@{}}{42.7\,($\pm$1.6)}\end{tabular}
& \begin{tabular}[c]{@{}c@{}}{13.6\%}\end{tabular}
\\ \midrule
\multirow{1}{*}[0.3em]{\textbf{DropTop}}                                                   
& \begin{tabular}[c]{@{}c@{}}{61.0\,($\pm$0.6)}\end{tabular}
& \begin{tabular}[c]{@{}c@{}}{---}\end{tabular}
& \begin{tabular}[c]{@{}c@{}}{37.6\,($\pm$1.9)}\end{tabular}
& \begin{tabular}[c]{@{}c@{}}{---}\end{tabular}
\\
\bottomrule
\end{tabular}
\vspace{-0.1cm}
\caption{Performance of ER+\algname{} with alternative design choices: the soft drop mask for attentive debiasing\,(Section 4.1) and a common drop intensity for all classes\,(Section 4.2).}
\label{tbl:other_design}

\end{table}

\subsection{Considerations of Alternative Design Choices}

\noindent{\textbf{Soft Drop Mask for Attentive Debiasing.}} We test a \textit{soft} drop mask by
replacing the \textit{hard} drop mask in Eq.~\eqref{eq:drop_mask}. Specifically, the soft drop mask linearly reduces the dependence on a shortcut feature as its attention value increases. That is, the soft drop mask at $(i,j)$ in a drop mask $M$ is determined by
\begin{equation}
M_{i,j} =
\begin{cases}
\frac{{\rm rank\text{-}}\kappa(i,j)}{|{\rm top\text{-}}\kappa(A_{fuse})|} & \text{if}\!\!\quad (i,j) \in {\rm top\text{-}}\kappa(A_{fuse}) \\ 
1 & \text{otherwise},\\ 
\end{cases}
\label{eq:alt_drop_mask}
\vspace*{-0.15cm}
\end{equation}
where ${\rm top\text{-}}\kappa(A_{fuse})$ returns the set of the top-$\kappa\%$ elements of $A_{fuse}$ and ${\rm rank\text{-}}\kappa(i,j)$ returns the rank of the value at $(i,j)$ within ${\rm top\text{-}}\kappa(A_{fuse})$ in descending order. Table \ref{tbl:other_design} shows a slight degradation by the soft drop mask, meaning that it is more effective to lessen reliance on shortcut features by outright excluding them.

\smallskip
\noindent{\textbf{Common Drop Intensity for All Classes.}} We test a \textit{common} drop intensity shared across classes by replacing class-wise drop intensities that capture diverse sensitivities of classes to shortcut features. Using a single common drop intensity notably degrades performance by 5.6\% and 13.6\% in terms of average accuracy and forgetting, as shown in Table \ref{tbl:other_design}. This result provides clear evidence for the necessity of maintaining class-wise drop intensities. 


\subsection{Ablation Studies for Unbiased Setup}

Table \ref{tbl:ablation_unbiased} compares \algname{} with its three variants across the unbiased datasets, Split ImageNet-OnlyFG and Split ImageNet-Stylized, where a sharper degradation of the performance indicates a stronger dependence on shortcut features. In general, the results for the unbiased datasets are consistent with those for the biased datasets. 

\def\arraystretch{0.7}
\begin{table}[t]
\scriptsize
\centering
\begin{tabular}[c]
{@{}l||c|c|c|cc@{}}
\toprule
\multicolumn{1}{c||}{\textbf{}} & \multicolumn{1}{c|}{\textbf{}} & \multicolumn{2}{c|}{\textbf{No Drop Int. Shift}} & \multicolumn{1}{c}{} \\
\multicolumn{1}{c||}{\hspace{-0.3cm}\textbf{Method}} 
& \multicolumn{1}{c|}{\textbf{No MF}\hspace{-1cm}}
& \multicolumn{1}{c|}{\textbf{Rand Drop}}
& \multicolumn{1}{c|}{\textbf{Fixed Drop}}
& \multicolumn{1}{c}{\textbf{DropTop}}

\\ \toprule




\multirow{1}{*}[0.4em]{ER\hspace{-0.15cm}}                                
& \begin{tabular}[c]{@{}c@{}}{35.2\hspace{-1cm}}\end{tabular}
& \begin{tabular}[c]{@{}c@{}}{38.1}\end{tabular}
& \begin{tabular}[c]{@{}c@{}}{39.1}\end{tabular}
& \begin{tabular}[c]{@{}c@{}}{\textbf{39.4}}\end{tabular}
\\

\multirow{1}{*}[0.4em]{DER++\hspace{-0.15cm}}                                                   
& \begin{tabular}[c]{@{}c@{}}{34.0\hspace{-1cm}}\end{tabular}
& \begin{tabular}[c]{@{}c@{}}{33.7}\end{tabular}
& \begin{tabular}[c]{@{}c@{}}{\textbf{38.0}}\end{tabular}
& \begin{tabular}[c]{@{}c@{}}{{37.1}}\end{tabular}
\\

\multirow{1}{*}[0.4em]{MIR\hspace{-0.15cm}}                                               
& \begin{tabular}[c]{@{}c@{}}{{33.1}\hspace{-1cm}}\end{tabular}
& \begin{tabular}[c]{@{}c@{}}{{32.3}}\end{tabular}
& \begin{tabular}[c]{@{}c@{}}{\textbf{36.6}}\end{tabular}
& \begin{tabular}[c]{@{}c@{}}{{35.8}}\end{tabular}
\\

\multirow{1}{*}[0.4em]{GSS\hspace{-0.15cm}}                                 
& \begin{tabular}[c]{@{}c@{}}{33.4\hspace{-1cm}}\end{tabular}
& \begin{tabular}[c]{@{}c@{}}{32.2}\end{tabular}
& \begin{tabular}[c]{@{}c@{}}{32.4}\end{tabular}
& \begin{tabular}[c]{@{}c@{}}{\textbf{36.2}}\end{tabular}
\\

\multirow{1}{*}[0.4em]{ASER\hspace{-0.15cm}}                                               
& \begin{tabular}[c]{@{}c@{}}{33.0\hspace{-1cm}}\end{tabular}
& \begin{tabular}[c]{@{}c@{}}{31.7}\end{tabular}
& \begin{tabular}[c]{@{}c@{}}{33.1}\end{tabular}
& \begin{tabular}[c]{@{}c@{}}{\textbf{33.4}}\end{tabular}
\\
\midrule

\multirow{1}{*}[0.4em]{\textbf{AVG}\,\textbf{(Degrad.)}\hspace{-0.15cm}}                                               
& \begin{tabular}[c]{@{}c@{}}{{33.7}\,(\textbf{-7.9\%})\hspace{-1cm}}\end{tabular}
& \begin{tabular}[c]{@{}c@{}}{{33.6}\,(\textbf{-8.3\%})}\end{tabular}
& \begin{tabular}[c]{@{}c@{}}{{35.8}\,(\textbf{-1.6\%})}\end{tabular}
& \begin{tabular}[c]{@{}c@{}}{\textbf{{36.4}}}\end{tabular}
\\
\bottomrule
\end{tabular}
\caption{Ablation study on multi-level feature fusion and drop intensity shifting with respect to the accuracy $A_{avg}$ averaged over the unbiased datasets: Split ImageNet-OnlyFG and Split ImageNet-Stylized. The highest values are marked in bold. }
\label{tbl:ablation_unbiased}

\end{table}

\smallskip
\noindent{\textbf{Multi-level Feature Fusion.}}
The first variant, \textit{No MF}, drops only high-level features without multi-level fusion. Compared with the results of the biased datasets, the degradation by No MF on the unbiased datasets is sharper
from the degradation of 4.9\% to 7.9\%, meaning that the multi-level fusion reduces undesirable bias towards shortcuts. Therefore, the multi-level fusion is essential to precisely capture shortcut features.

\begin{table}[!t]
\small
\centering
\renewcommand{\arraystretch}{0.5}
\resizebox{0.9\columnwidth}{!}{%
\begin{tabular}[c]
{@{}c|llc@{}}
\toprule
\multicolumn{1}{c|}{} & 
\multicolumn{1}{c}{\textbf{GPU Memory}\,(GB)} & 
\multicolumn{1}{c}{\hspace*{0.4cm}\textbf{Running Time}\,(mins.)} &
\\
\toprule
\vspace*{0.05cm}
\begin{tabular}[c]{@{}c@{}}\textbf{ER}\end{tabular}       
& \begin{tabular}[c]{@{}c@{}}\hspace*{0.4cm}4.509\end{tabular}
& \begin{tabular}[c]{@{}c@{}}\hspace*{0.4cm}42.55\end{tabular} 
\\
\begin{tabular}[c]{@{}c@{}}\textbf{ER+DropTop}\end{tabular}       
& \begin{tabular}[c]{@{}c@{}}\hspace*{0.4cm}4.563\,(+0.054\,/\,1.183\%)\end{tabular}
& \begin{tabular}[c]{@{}c@{}}\hspace*{0.4cm}49.26\,(+6.71\,/\,13.62\%)\end{tabular} 
\vspace*{-0.05cm}
\\  
\bottomrule

\end{tabular}}
\vspace{-0.1cm}
\caption{Computational overhead incurred by DropTop.}
\label{tbl:complexity}
\end{table}

\smallskip
\noindent{\textbf{Adaptive Intensity Shifting.}} 
\textit{Rand Drop} drops randomly chosen features, and \textit{Fixed Drop} drops the features with the highest attention scores, where neither of them uses adaptive intensity shifting. As a result, we consistently observe that using adaptive intensity shifting makes further improvements over the two variants by 8.3\% and 1.6\%. In particular, Rand Drop becomes the worst for the unbiased setup, emphasizing that random drop-based regularization is not suitable for debiasing the shortcut features.

\subsection{Computational Complexity}
Table \ref{tbl:complexity} reports the GPU memory and OCL time of ER \emph{with and without} DropTop on Split ImageNet-9, showing that DropTop \emph{efficiently} debiases shortcuts. The additional cost of DropTop is only $\frac{p+1}{p}$ forward passes on average per iteration, where $p$ is the alternating period ($p\!=\!3$ in the paper); it is much smaller than the total cost of ER (especially including backward passes). A forward pass is inexpensive, costing roughly one-third as much as a backward pass\,\cite{li13pytorch}.

\def\arraystretch{0.7}
\begin{table}[t]
\scriptsize
\centering
\resizebox{1.0\columnwidth}{!}{%
\begin{tabular}[c]
{@{}l|ccc|cccc@{}}
\toprule
\multicolumn{1}{c|}{\textbf{}}& \multicolumn{3}{c}{\textbf{Split CIFAR-100}} & \multicolumn{3}{c}{\textbf{Split CIFAR-10}} \\
\multicolumn{1}{c|}{\textbf{Method}}
& \multicolumn{1}{c}{\hspace{-0.07cm}\bm{$A_{last}$}\,($\uparrow$)}
& \multicolumn{1}{c}{\hspace{-0.07cm}\bm{$A_{avg}$}\,($\uparrow$)}
& \multicolumn{1}{c|}{\hspace{-0.07cm}\bm{$F_{last}$}\,($\downarrow$)}
& \multicolumn{1}{c}{\hspace{-0.07cm}\bm{$A_{last}$}\,($\uparrow$)}
& \multicolumn{1}{c}{\hspace{-0.07cm}\bm{$A_{avg}$}\,($\uparrow$)}
& \multicolumn{1}{c}{\hspace{-0.07cm}\bm{$F_{last}$}\,($\downarrow$)}
\\ \toprule
\multirow{1}{*}[0.4em]{L2P\!\!\!}                                
& \begin{tabular}[c]{@{}c@{}}{ \hspace{-0.25cm}{72.3}\,($\pm$0.6) }\end{tabular}
& \begin{tabular}[c]{@{}c@{}}{ \hspace{-0.25cm}{81.5}\,($\pm$1.2) }\end{tabular}
& \begin{tabular}[c]{@{}c@{}}{ \hspace{-0.25cm}{24.3}\,($\pm$0.9) }\end{tabular}
& \begin{tabular}[c]{@{}c@{}}{ \hspace{-0.25cm}{85.8}\,($\pm$1.1) }\end{tabular}
& \begin{tabular}[c]{@{}c@{}}{ \hspace{-0.25cm}{92.2}\,($\pm$1.3) }\end{tabular}
& \begin{tabular}[c]{@{}c@{}}{ \hspace{-0.25cm}{16.6}\,($\pm$1.4) }\end{tabular}
\\
\multirow{1}{*}[0.4em]{\textbf{~~$+$\algname{}}\!\!\!}  
& \begin{tabular}[c]{@{}c@{}}{ \hspace{-0.25cm}\textbf{74.3}\,($\pm$0.7) }\end{tabular}
& \begin{tabular}[c]{@{}c@{}}{ \hspace{-0.25cm}\textbf{82.1}\,($\pm$0.6) }\end{tabular}
& \begin{tabular}[c]{@{}c@{}}{ \hspace{-0.25cm}\textbf{21.9}\,($\pm$1.1) }\end{tabular}
& \begin{tabular}[c]{@{}c@{}}{ \hspace{-0.25cm}\textbf{89.3}\,($\pm$0.7) }\end{tabular}
& \begin{tabular}[c]{@{}c@{}}{ \hspace{-0.25cm}\textbf{94.0}\,($\pm$0.2) }\end{tabular}
& \begin{tabular}[c]{@{}c@{}}{ \hspace{-0.25cm}\textbf{12.2}\,($\pm$0.8) }\end{tabular}
\\ \addlinespace[0.3ex]\cdashline{1-8}\addlinespace[0.5ex]
\multirow{1}{*}[0.4em]{~~Rel. Improv.\!\!\!}  
& \begin{tabular}[c]{@{}c@{}}{ 2.7\% }\end{tabular}
& \begin{tabular}[c]{@{}c@{}}{ 0.7\% }\end{tabular}
& \begin{tabular}[c]{@{}c@{}}{ 9.9\% }\end{tabular}
& \begin{tabular}[c]{@{}c@{}}{ 4.0\% }\end{tabular}
& \begin{tabular}[c]{@{}c@{}}{ 1.9\% }\end{tabular}
& \begin{tabular}[c]{@{}c@{}}{ 26.7\% }\end{tabular}
\\
\midrule

\multirow{1}{*}[0.4em]{DualPrompt\!\!\!}                                                         
& \begin{tabular}[c]{@{}c@{}}{ \hspace{-0.25cm}{73.5}\,($\pm$0.5) }\end{tabular}
& \begin{tabular}[c]{@{}c@{}}{ \hspace{-0.25cm}\textbf{83.2}\,($\pm$0.8) }\end{tabular}
& \begin{tabular}[c]{@{}c@{}}{ \hspace{-0.25cm}{22.5}\,($\pm$0.8) }\end{tabular}
& \begin{tabular}[c]{@{}c@{}}{ \hspace{-0.25cm}{89.5}\,($\pm$1.6) }\end{tabular}
& \begin{tabular}[c]{@{}c@{}}{ \hspace{-0.25cm}{94.9}\,($\pm$1.3) }\end{tabular}
& \begin{tabular}[c]{@{}c@{}}{ \hspace{-0.25cm}{11.8}\,($\pm$0.7) }\end{tabular}
\\
\multirow{1}{*}[0.4em]{\textbf{~~$+$\algname{}}\!\!\!}   
& \begin{tabular}[c]{@{}c@{}}{ \hspace{-0.25cm}\textbf{74.0}\,($\pm$0.4) }\end{tabular}
& \begin{tabular}[c]{@{}c@{}}{ \hspace{-0.25cm}{82.7}\,($\pm$0.4) }\end{tabular}
& \begin{tabular}[c]{@{}c@{}}{ \hspace{-0.25cm}\textbf{21.5}\,($\pm$0.6) }\end{tabular}
& \begin{tabular}[c]{@{}c@{}}{ \hspace{-0.25cm}\textbf{90.5}\,($\pm$0.4) }\end{tabular}
& \begin{tabular}[c]{@{}c@{}}{ \hspace{-0.25cm}\textbf{95.0}\,($\pm$0.2) }\end{tabular}
& \begin{tabular}[c]{@{}c@{}}{ \hspace{-0.25cm}\textbf{10.5}\,($\pm$0.5) }\end{tabular}
\\ \addlinespace[0.3ex]\cdashline{1-8}\addlinespace[0.5ex]
\multirow{1}{*}[0.4em]{~~Rel. Improv.\!\!\!}  
& \begin{tabular}[c]{@{}c@{}}{ 0.7\% }\end{tabular}
& \begin{tabular}[c]{@{}c@{}}{ -0.6\% }\end{tabular}
& \begin{tabular}[c]{@{}c@{}}{ 4.5\% }\end{tabular}
& \begin{tabular}[c]{@{}c@{}}{ 1.2\% }\end{tabular}
& \begin{tabular}[c]{@{}c@{}}{ 0.1\% }\end{tabular}
& \begin{tabular}[c]{@{}c@{}}{ 11.2\% }\end{tabular}
\\
\bottomrule
\end{tabular}
}
\caption{Performance of DropTop on top of pretrained ViT-based CL algorithms, L2P\,\cite{wang2022learning} and DualPrompt\,\cite{wang2022dualprompt}, on Split CIFAR-100, and Split CIFAR-10. The highest values are marked in bold.}
\label{tbl:ptch_cifars}
\end{table}

\begin{figure}[!t]
    \subfloat[Debiasing the \textbf{background} shortcut bias.]{%
      \includegraphics[clip,width=0.98\columnwidth]{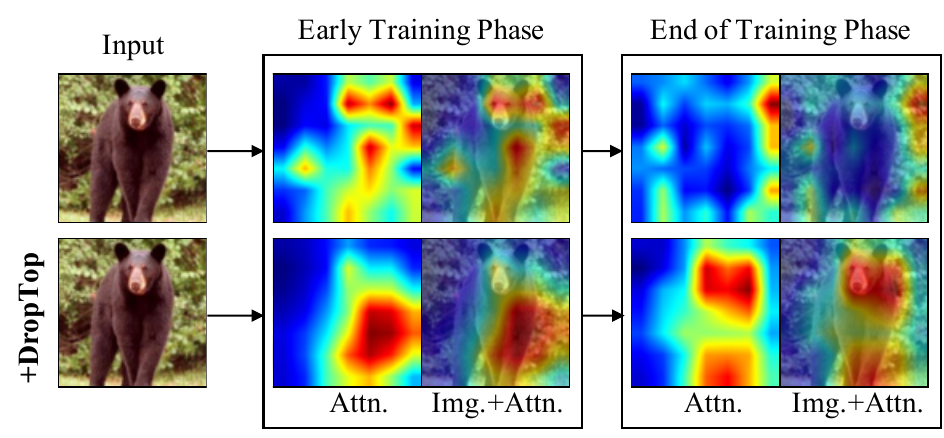}%
    }
    \vspace*{0.1cm}
    \subfloat[Debiasing the \textbf{local cue} bias.]{%
      \includegraphics[clip,width=0.98\columnwidth]{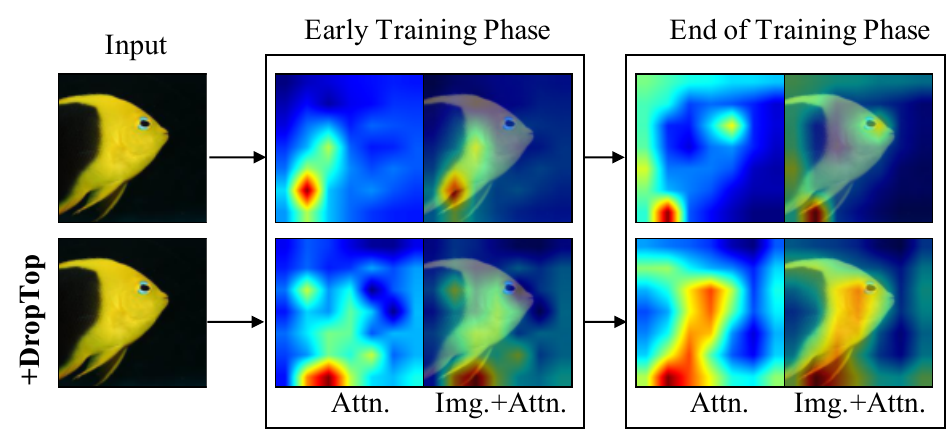}%
    }
    \caption{Additional visualization of the gradual debiasing process by \algname{}. The results are the activation maps from ER and ER$+$\algname{} trained on Split ImageNet-9: (a) and (b) are respectively related to debiasing the background and local cue bias during training, where the bias is the reliance on the grass background in (a) and on a local part of a fish\,(e.g., its eye without the body) in (b). 
    }
    \label{fig:activation_maps_add}
\end{figure}

\section{F \quad Debiasing Pretrained ViT-based CL}

In addition to Table \ref{tbl:perf_dualprompt}, Table \ref{tbl:ptch_cifars} shows the performance of L2P\,\cite{wang2022learning} and DualPrompt\,\cite{wang2022dualprompt} with and without \algname{} on Split CIFAR-100 and Split CIFAR-10 in an online setting. Overall, the results for the CIFAR datasets are consistent with those for the ImageNet datasets. Specifically, \algname{} improves $F_{last}$, $A_{last}$, and $A_{avg}$ by 13.1\%, 2.2\%, and 0.5\%, respectively, on average across the datasets and algorithms. 
The relatively modest improvement observed in $A_{avg}$ can be attributed to the slow convergence of the prompt tuning mechanism, which has been widely acknowledged\,\cite{HuangQWYSLL22, su2022transferability}, in L2P and DualPrompt. This slow convergence leads to a gradually increasing dependence on shortcut features, which reduces the effectiveness of debiasing in the early stages of training. Consequently, the improvement observed in $A_{avg}$ becomes less significant than that observed in $A_{last}$.
Nevertheless, the overall improvements by \algname{} again emphasize the compelling need for mitigating the shortcut bias, irrespective of backbone networks.

\section{G \quad Limitations and Future Work}

\algname{} has consistently improved performance on a range of datasets when combined with different OCL algorithms. However, it is necessary to formally formulate the effectiveness of \algname{} in light of the characteristics of datasets and OCL algorithms. Although debiasing shortcut features by \algname{} is effective since DNNs are highly susceptible to the shortcut bias\,\cite{shah2020pitfalls}, as shown in Section Evaluation, its effectiveness varies by OCL algorithms and datasets. Therefore, determining the effectiveness of debiasing shortcuts based on such characteristics would be an interesting research area.

\section{H \quad Additional Visualization of Debiasing}


Last, in addition to Figure \ref{fig:activation_maps}, we present another visualization in Figure \ref{fig:activation_maps_add} which was omitted owing to lack of space. Again, Figure \ref{fig:activation_maps_add} verifies that \algname{} gradually alleviates the shortcut bias as the training proceeds whereas the original ER increasingly relies on the shortcut bias---the background and local cue.

\end{document}